\documentclass[10pt,twocolumn,letterpaper]{article}

\usepackage{iccv}
\usepackage{times}
\usepackage{epsfig}
\usepackage{graphicx}
\usepackage{amsmath}
\usepackage{amssymb}
\usepackage{caption}
\usepackage[accsupp]{axessibility}
\usepackage[table,xcdraw]{xcolor}

\usepackage[pagebackref=true,breaklinks=true,letterpaper=true,colorlinks,bookmarks=false]{hyperref}

\iccvfinalcopy 

\def\iccvPaperID{3364} 

\ificcvfinal\fi

\usepackage{booktabs}
\usepackage[export]{adjustbox}
\usepackage{tikz}
\usetikzlibrary{spy,calc}
\usepackage{subcaption}

\usepackage{mathtools}

\DeclarePairedDelimiter\floor{\lfloor}{\rfloor}

\usepackage[capitalize]{cleveref}
\crefname{section}{Sec.}{Secs.}
\Crefname{section}{Section}{Sections}
\Crefname{table}{Table}{Tables}
\crefname{table}{Tab.}{Tabs.}

\definecolor{turquoise}{cmyk}{0.65,0,0.1,0.3}
\definecolor{purple}{rgb}{0.65,0,0.65}
\definecolor{darkgreen}{rgb}{0, 0.5, 0}
\definecolor{orange}{rgb}{0.8, 0.6, 0.2}
\definecolor{darkred}{rgb}{0.6, 0.1, 0.05}
\definecolor{blueish}{rgb}{0.0, 0.3, .6}
\definecolor{lightgray}{rgb}{0.7, 0.7, .7}
\definecolor{pink}{rgb}{1, 0, 1}
\definecolor{greyblue}{rgb}{0.25, 0.25, 1}

\newcommand{\todo}[1]{\textcolor{darkred}{[TODO: #1]}}

\newcommand{\hb}{\hat{\boldsymbol{h}}}
\newcommand{\omegai}{\hat{\boldsymbol{\omega}}_i}
\newcommand{\omegao}{\hat{\boldsymbol{\omega}}_o}

\newcommand{\omegan}{\hat{\boldsymbol{\omega}}_i^n}
\newcommand{\feat}{\boldsymbol{x}}
\newcommand{\bp}{\mathbf{p}}

\newcommand{\uvn}{\hat{\mathbf{n}}}

\newcommand{\density}{\sigma}
\newcommand{\ray}{\vec{\mathbf{r}}}
\newcommand{\modelname}{Neural Microfacet Fields}
\newcommand{\bsdf}{BRDF}
\newcommand{\point}{\bp}

\begin{document}

\title{\modelname{} for Inverse Rendering}

\author{Alexander Mai\\
{\normalsize UC San Diego} \\
{\tt\small atm008@ucsd.edu}
\and
Dor Verbin\\
{\normalsize Google Research}\\
{\tt\small dorverbin@google.com}
\and
Falko Kuester\\
{\normalsize UC San Diego}\\
{\tt\small fkuester@ucsd.edu}
\and
Sara Fridovich-Keil\\
{\normalsize UC Berkeley}\\
{\tt\small sfk@eecs.berkeley.edu}
}

\newcommand{\plotall}[1]{%
  \adjincludegraphics[trim={{0\width} {0\height} {0\width} {0\height}}, clip, height=0.22\linewidth]{#1}%
}

\newcommand{\plotfourtrim}[1]{%
  \adjincludegraphics[trim={{0\width} {0.02\height} {0\width} {0.02\height}}, clip, width=0.24\linewidth]{#1}%
}
\newcommand{\plotfour}[1]{%
  \adjincludegraphics[trim={{0\width} {0.0\height} {0\width} {0.0\height}}, clip, width=0.24\linewidth]{#1}%
}
\newcommand{\plotscene}[1]{
\begin{figure*}[h]
\begin{tabular}{l@{~~}c@{}c@{}c@{}c@{}}
\rotatebox{90}{~~~~~~~~~~~Novel View} & \plotfourtrim{images/#1/gt_final.png} & \plotfourtrim{images/#1/final.png} & \plotfourtrim{nvdiffrec/#1/final.png} & \plotfourtrim{nvdiffrecmc/#1/final.png} \\
\rotatebox{90}{~~~~~~~~~~~~~~Normals} & \plotfourtrim{images/#1/gt_normals.png} & \plotfourtrim{images/#1/normals.png} & \plotfourtrim{nvdiffrec/#1/normals.png} & \plotfourtrim{nvdiffrecmc/#1/normals.png} \\
\rotatebox{90}{Environment} & \plotfour{images/#1/gt_pano.png} & \plotfour{images/#1/mapped_pano.png} & \plotfour{nvdiffrec/#1/mapped_pano.png} & \plotfour{nvdiffrecmc/#1/mapped_pano.png} \\
& Ground Truth & Ours & NVDiffRec & NVDiffRecMC
\end{tabular}
\caption{\textbf{Results on the \emph{#1} scene}, compared to NVDiffRec \cite{munkberg2021nvdiffrec} and NVDiffRecMC \cite{hasselgren2022nvdiffrecmc}.}
\label{fig:#1}
\end{figure*}
}

\newcommand{\plotfourtrimtop}[1]{%
  \adjincludegraphics[trim={{0\width} {0.05\height} {0\width} {0.2\height}}, clip, width=0.24\linewidth]{#1}%
}

\newcommand{\plotship}[1]{
\begin{figure*}[h]
\begin{tabular}{l@{~~}c@{}c@{}c@{}c@{}}
\rotatebox{90}{~~~~~~~~~~~Novel View} & \plotfourtrimtop{images/#1/gt_final.png} & \plotfourtrimtop{images/#1/final.png} & \plotfourtrimtop{nvdiffrec/#1/final.png} & \plotfourtrimtop{nvdiffrecmc/#1/final.png} \\
\rotatebox{90}{~~~~~~~~~~~~~~Normals} & \plotfourtrimtop{images/#1/gt_normals.png} & \plotfourtrimtop{images/#1/normals.png} & \plotfourtrimtop{nvdiffrec/#1/normals.png} & \plotfourtrimtop{nvdiffrecmc/#1/normals.png} \\
\rotatebox{90}{Environment} & \plotfour{images/#1/gt_pano.png} & \plotfour{images/#1/mapped_pano.png} & \plotfour{nvdiffrec/#1/mapped_pano.png} & \plotfour{nvdiffrecmc/#1/mapped_pano.png} \\
& Ground Truth & Ours & NVDiffRec & NVDiffRecMC
\end{tabular}
\caption{\textbf{Results on the \emph{#1} scene}, compared to NVDiffRec \cite{munkberg2021nvdiffrec} and NVDiffRecMC \cite{hasselgren2022nvdiffrecmc}. Since our method, NVDiffRec, and NVDiffRecMC do not model refraction, they are not able to handle the water well.}
\label{fig:#1}
\end{figure*}
}

\newcommand{\plotfourtrimmore}[1]{%
  \adjincludegraphics[trim={{0\width} {0.2\height} {0\width} {0.3\height}}, clip, width=0.24\linewidth]{#1}%
}
\newcommand{\plotsceneshort}[1]{
\begin{figure*}[h]
\begin{tabular}{l@{~~}c@{}c@{}c@{}c@{}}
\rotatebox{90}{~~~~Novel View} & \plotfourtrimmore{images/#1/gt_final.png} & \plotfourtrimmore{images/#1/final.png} & \plotfourtrimmore{nvdiffrec/#1/final.png} & \plotfourtrimmore{nvdiffrecmc/#1/final.png} \\
\rotatebox{90}{~~~~~Normals} & \plotfourtrimmore{images/#1/gt_normals.png} & \plotfourtrimmore{images/#1/normals.png} & \plotfourtrimmore{nvdiffrec/#1/normals.png} & \plotfourtrimmore{nvdiffrecmc/#1/normals.png} \\
\rotatebox{90}{Environment} & \plotfour{images/#1/gt_pano.png} & \plotfour{images/#1/mapped_pano.png} & \plotfour{nvdiffrec/#1/mapped_pano.png} & \plotfour{nvdiffrecmc/#1/mapped_pano.png} \\
& Ground Truth & Ours & NVDiffRec & NVDiffRecMC
\end{tabular}
\caption{\textbf{Results on the \emph{#1} scene}, compared to NVDiffRec \cite{munkberg2021nvdiffrec} and NVDiffRecMC \cite{hasselgren2022nvdiffrecmc}.}
\label{fig:#1}
\end{figure*}
}

\newcommand{\plotscenebottom}[1]{
\begin{figure*}[h]
\begin{tabular}{l@{~~}c@{}c@{}c@{}c@{}}
\rotatebox{90}{~~~~Novel View} & \plotfourtrimtop{images/#1/gt_final.png} & \plotfourtrimtop{images/#1/final.png} & \plotfourtrimtop{nvdiffrec/#1/final.png} & \plotfourtrimtop{nvdiffrecmc/#1/final.png} \\
\rotatebox{90}{~~~~~Normals} & \plotfourtrimtop{images/#1/gt_normals.png} & \plotfourtrimtop{images/#1/normals.png} & \plotfourtrimtop{nvdiffrec/#1/normals.png} & \plotfourtrimtop{nvdiffrecmc/#1/normals.png} \\
\rotatebox{90}{Environment} & \plotfour{images/#1/gt_pano.png} & \plotfour{images/#1/mapped_pano.png} & \plotfour{nvdiffrec/#1/mapped_pano.png} & \plotfour{nvdiffrecmc/#1/mapped_pano.png} \\
& Ground Truth & Ours & NVDiffRec & NVDiffRecMC
\end{tabular}
\caption{\textbf{Results on the \emph{#1} scene}, compared to NVDiffRec \cite{munkberg2021nvdiffrec} and NVDiffRecMC \cite{hasselgren2022nvdiffrecmc}.}
\label{fig:#1}
\end{figure*}
}

\newcommand{\plotfourtrimbottom}[1]{%
  \adjincludegraphics[trim={{0\width} {0.2\height} {0\width} {0.05\height}}, clip, width=0.24\linewidth]{#1}%
}
\newcommand{\plotscenetop}[1]{
\begin{figure*}[h]
\begin{tabular}{l@{~~}c@{}c@{}c@{}c@{}}
\rotatebox{90}{~~~~Novel View} & \plotfourtrimbottom{images/#1/gt_final.png} & \plotfourtrimbottom{images/#1/final.png} & \plotfourtrimbottom{nvdiffrec/#1/final.png} & \plotfourtrimbottom{nvdiffrecmc/#1/final.png} \\
\rotatebox{90}{~~~~~Normals} & \plotfourtrimbottom{images/#1/gt_normals.png} & \plotfourtrimbottom{images/#1/normals.png} & \plotfourtrimbottom{nvdiffrec/#1/normals.png} & \plotfourtrimbottom{nvdiffrecmc/#1/normals.png} \\
\rotatebox{90}{Environment} & \plotfour{images/#1/gt_pano.png} & \plotfour{images/#1/mapped_pano.png} & \plotfour{nvdiffrec/#1/mapped_pano.png} & \plotfour{nvdiffrecmc/#1/mapped_pano.png} \\
& Ground Truth & Ours & NVDiffRec & NVDiffRecMC
\end{tabular}
\caption{\textbf{Results on the \emph{#1} scene}, compared to NVDiffRec \cite{munkberg2021nvdiffrec} and NVDiffRecMC \cite{hasselgren2022nvdiffrecmc}.}
\label{fig:#1}
\end{figure*}
}

\newcommand{\plottraining}[1]{
\begin{figure*}[h]
\centering
\begin{tabular}{c@{~}c@{~}c@{~}c}
\plotfour{#1_overtime/im100.png}
& \plotfour{#1_overtime/im500.png}
& \plotfour{#1_overtime/im900.png}
& \plotfour{#1_overtime/im30k.png} \\
\plotfour{#1_overtime/pano100.png}
& \plotfour{#1_overtime/pano500.png}
& \plotfour{#1_overtime/pano900.png}
& \plotfour{#1_overtime/pano30k.png} \\
100 steps & 500 steps & 900 steps & 30000 steps
\end{tabular}
\caption{\textbf{Snapshots of the \emph{#1} scene during optimization.} Early in training the object geometry is cloudy and the environment map is uniform, but as training proceeds the object develops a sharp surface and the environment map converges.}
\label{fig:#1 over time}
\end{figure*}
}

\newcommand{\plotlefttrim}[1]{%
  \adjincludegraphics[trim={{0\width} {0.06\height} {0\width} {0.06\height}}, clip, width=0.8\textwidth]{#1}%
}
\newcommand{\plotleft}[1]{%
  \adjincludegraphics[trim={{0\width} {0.0\height} {0\width} {0.0\height}}, clip, width=0.8\textwidth]{#1}%
}
\newcommand{\plotrighttrim}[1]{%
  \adjincludegraphics[trim={{0.03\width} {0.05\height} {0\width} {0.138\height}}, clip, width=\textwidth]{#1}%
}
\newcommand{\plotenvswap}{
\begin{figure}[h]
\centering
\begin{minipage}{0.39\linewidth}
\centering
\begin{tabular}{c@{~}c@{}}
\plotlefttrim{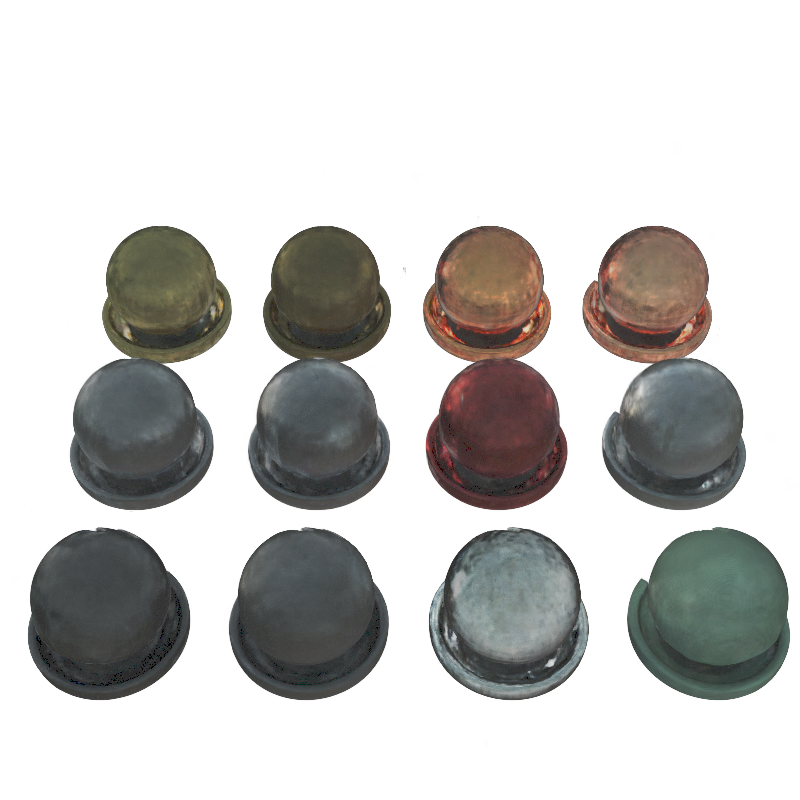} & \rotatebox{90}{~~~~~~~~~~$\downarrow$} \\
\emph{Materials} \bsdf{} & \\
\plotleft{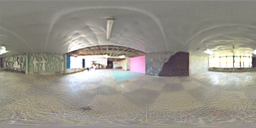} & \rotatebox{90}{~~~~~~$\downarrow$} \\
\emph{Helmet} Lighting \\
\end{tabular}
\end{minipage} 
\begin{minipage}{0.59\linewidth}
\centering
\begin{tabular}{c@{}}
\plotrighttrim{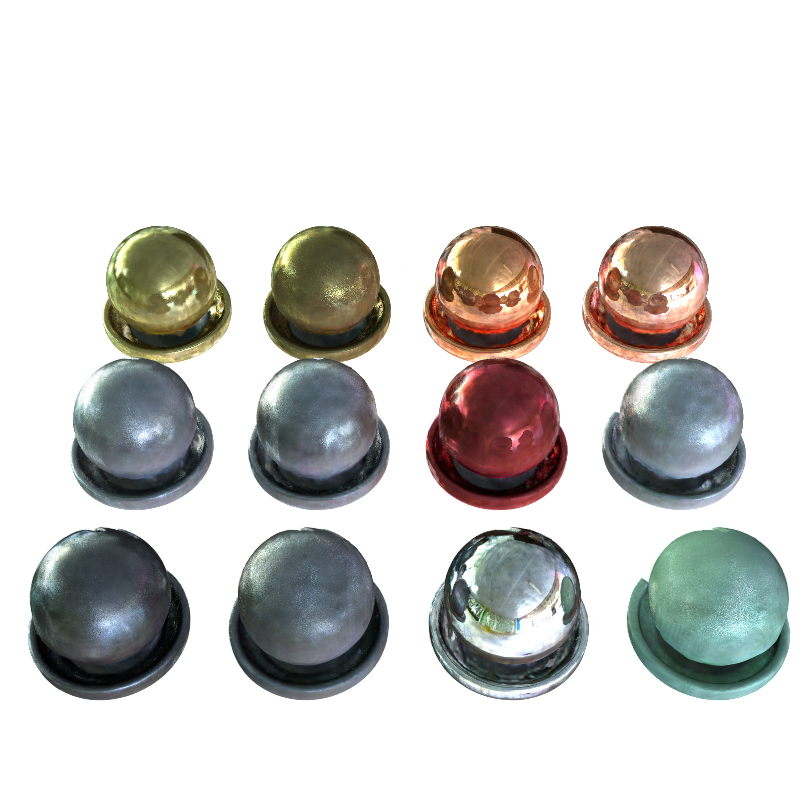} \\
Combined Rendering (a)
\end{tabular}
\end{minipage}
\includegraphics[width=\linewidth]{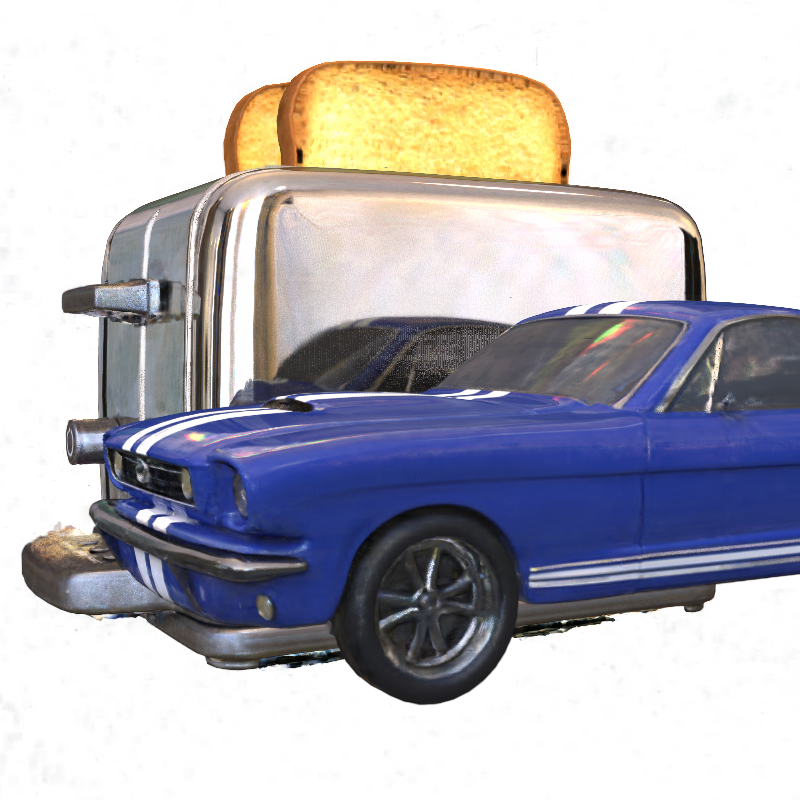}
Combined Rendering (b)

\caption{\textbf{Rendering with different illumination.} (a) shows the geometry and \bsdf{} (shown integrated against uniform white lighting) recovered from the \emph{materials} scene, rendered with the environment map recovered from the \emph{helmet} scene. (b) shows the geometry and \bsdf{} recovered from the \emph{toaster} and \emph{car} scene composed under the environment map recovered from the \emph{toaster} scene.}
\label{fig:environment swap}
\end{figure}
}

\newif\ifblackandwhitecycle
\gdef\patternnumber{0}

\pgfkeys{/tikz/.cd,
    zoombox paths/.style={
        draw=orange,
        very thick
    },
    black and white/.is choice,
    black and white/.default=static,
    black and white/static/.style={
        draw=white,
        zoombox paths/.append style={
            draw=white,
            postaction={
                draw=black,
                loosely dashed
            }
        }
    },
    black and white/static/.code={
        \gdef\patternnumber{1}
    },
    black and white/cycle/.code={
        \blackandwhitecycletrue
        \gdef\patternnumber{1}
    },
    black and white pattern/.is choice,
    black and white pattern/0/.style={},
    black and white pattern/1/.style={
            draw=white,
            postaction={
                draw=black,
                dash pattern=on 2pt off 2pt
            }
    },
    black and white pattern/2/.style={
            draw=white,
            postaction={
                draw=black,
                dash pattern=on 4pt off 4pt
            }
    },
    black and white pattern/3/.style={
            draw=white,
            postaction={
                draw=black,
                dash pattern=on 4pt off 4pt on 1pt off 4pt
            }
    },
    black and white pattern/4/.style={
            draw=white,
            postaction={
                draw=black,
                dash pattern=on 4pt off 2pt on 2 pt off 2pt on 2 pt off 2pt
            }
    },
    zoomboxarray inner gap/.initial=5pt,
    zoomboxarray columns/.initial=2,
    zoomboxarray rows/.initial=1,
    zoomboxarray heightmultiplier/.initial=0.5,
    subfigurename/.initial={},
    figurename/.initial={zoombox},
    zoomboxarray/.style={
        execute at begin picture={
            \begin{scope}[
                spy using outlines={%
                    zoombox paths,
                    width=\imagewidth / \pgfkeysvalueof{/tikz/zoomboxarray columns} - (\pgfkeysvalueof{/tikz/zoomboxarray columns} - 1) / \pgfkeysvalueof{/tikz/zoomboxarray columns} * \pgfkeysvalueof{/tikz/zoomboxarray inner gap} -\pgflinewidth,
                    height=\pgfkeysvalueof{/tikz/zoomboxarray heightmultiplier} * (\imageheight / \pgfkeysvalueof{/tikz/zoomboxarray rows} - (\pgfkeysvalueof{/tikz/zoomboxarray rows} - 1) / \pgfkeysvalueof{/tikz/zoomboxarray rows} * \pgfkeysvalueof{/tikz/zoomboxarray inner gap}-\pgflinewidth),
                    magnification=3,
                    every spy on node/.style={
                        zoombox paths
                    },
                    every spy in node/.style={
                        zoombox paths
                    }
                }
            ]
        },
        execute at end picture={
            \end{scope}
     \gdef\patternnumber{0}
        },
        spymargin/.initial=0.5em,
        zoomboxes xshift/.initial=1,
        zoomboxes right/.code=\pgfkeys{/tikz/zoomboxes xshift=1},
        zoomboxes left/.code=\pgfkeys{/tikz/zoomboxes xshift=-1},
        zoomboxes yshift/.initial=0,
        zoomboxes above/.code={
            \pgfkeys{/tikz/zoomboxes yshift=1},
            \pgfkeys{/tikz/zoomboxes xshift=0}
        },
        zoomboxes below/.code={
            \pgfkeys{/tikz/zoomboxes yshift=-1},
            \pgfkeys{/tikz/zoomboxes xshift=0}
        },
        caption margin/.initial=0ex, %
    },
    adjust caption spacing/.code={},
    image container/.style={
        inner sep=0pt,
        at=(image.north),
        anchor=north,
        adjust caption spacing
    },
    zoomboxes container/.style={
        inner sep=0pt,
        at=(image.north),
        anchor=north,
        name=zoomboxes container,
        xshift=\pgfkeysvalueof{/tikz/zoomboxes xshift}*(\imagewidth+\pgfkeysvalueof{/tikz/spymargin}),
        yshift=\pgfkeysvalueof{/tikz/zoomboxes yshift}*(\imageheight+\pgfkeysvalueof{/tikz/spymargin}+\pgfkeysvalueof{/tikz/caption margin}),
        adjust caption spacing
    },
    calculate dimensions/.code={
        \pgfpointdiff{\pgfpointanchor{image}{south west} }{\pgfpointanchor{image}{north east} }
        \pgfgetlastxy{\imagewidth}{\imageheight}
        \global\let\imagewidth=\imagewidth
        \global\let\imageheight=\imageheight
        \gdef\columncount{1}
        \gdef\rowcount{1}
        \gdef\zoomboxcount{1}
    },
    image node/.style={
        inner sep=0pt,
        name=image,
        anchor=south west,
        append after command={
            [calculate dimensions]
            node [image container,subfigurename=\pgfkeysvalueof{/tikz/figurename}-image] {\phantomimage}
            node [zoomboxes container,subfigurename=\pgfkeysvalueof{/tikz/figurename}-zoom] {\phantomimage}
        }
    },
    color code/.style={
        zoombox paths/.append style={draw=#1}
    },
    connect zoomboxes/.style={
    spy connection path={\draw[draw=none,zoombox paths] (tikzspyonnode) -- (tikzspyinnode);}
    },
    help grid code/.code={
        \begin{scope}[
                x={(image.south east)},
                y={(image.north west)},
                font=\footnotesize,
                help lines,
                overlay
            ]
            \foreach \x in {0,1,...,9} {
                \draw(\x/10,0) -- (\x/10,1);
                \node [anchor=north] at (\x/10,0) {0.\x};
            }
            \foreach \y in {0,1,...,9} {
                \draw(0,\y/10) -- (1,\y/10);                        \node [anchor=east] at (0,\y/10) {0.\y};
            }
        \end{scope}
    },
    help grid/.style={
        append after command={
            [help grid code]
        }
    },
}

\newcommand\phantomimage{%
    \phantom{%
        \rule{\imagewidth}{\imageheight}%
    }%
}
\newcommand\zoombox[2][]{
    \begin{scope}[zoombox paths]
        \pgfmathsetmacro\xpos{
            (\columncount-1)*(\imagewidth / \pgfkeysvalueof{/tikz/zoomboxarray columns} + \pgfkeysvalueof{/tikz/zoomboxarray inner gap} / \pgfkeysvalueof{/tikz/zoomboxarray columns} ) + \pgflinewidth
        }
        \pgfmathsetmacro\ypos{
            (\rowcount-1) * (\imageheight / \pgfkeysvalueof{/tikz/zoomboxarray rows} + \pgfkeysvalueof{/tikz/zoomboxarray inner gap} / \pgfkeysvalueof{/tikz/zoomboxarray rows} ) + 0.5*\pgflinewidth
        }
        \edef\dospy{\noexpand\spy [
            #1,
            zoombox paths/.append style={
                black and white pattern=\patternnumber
            },
            every spy on node/.append style={#1},
            x=\imagewidth,
            y=\imageheight
        ] on (#2) in node [anchor=north west] at ($(zoomboxes container.north west)+(\xpos pt,-\ypos pt)$);}
        \dospy
        \pgfmathtruncatemacro\pgfmathresult{ifthenelse(\columncount==\pgfkeysvalueof{/tikz/zoomboxarray columns},\rowcount+1,\rowcount)}
        \global\let\rowcount=\pgfmathresult
        \pgfmathtruncatemacro\pgfmathresult{ifthenelse(\columncount==\pgfkeysvalueof{/tikz/zoomboxarray columns},1,\columncount+1)}
        \global\let\columncount=\pgfmathresult
        \ifblackandwhitecycle
            \pgfmathtruncatemacro{\newpatternnumber}{\patternnumber+1}
            \global\edef\patternnumber{\newpatternnumber}
        \fi
    \end{scope}
}

\newcommand{\plottrim}[1]{%
  \adjincludegraphics[trim={{0\width} {0.2\height} {0\width} {0.2\height}}, clip, width=0.24\linewidth]{#1}%
}

\definecolor{col1}{HTML}{e0d291}
\definecolor{col2}{HTML}{d66079}
\newcommand\plotzoomed[1]{
    \raisebox{-0.22\height}{
    \begin{tikzpicture}[
    zoomboxarray,
    zoomboxes below,
    connect zoomboxes,
    zoombox paths/.append style={thick}]
        \node[image node]{\plottrim{#1}};
        \zoombox[magnification=4,color code=col1]{0.28,0.670}
        \zoombox[magnification=4,color code=col2]{0.85,0.35}
    \end{tikzpicture}
    }
}

\newcommand{\plotteaser}[1]{
\twocolumn[{
\renewcommand\twocolumn[1][]{#1}
\maketitle
\begin{center}
\begin{tabular}{l@{~}c@{}c@{}c@{}} \vspace{-1.3cm}
\rotatebox{90}{~~~~~~~~~~~~~~~~Ours}
& \plotzoomed{images/#1/final.png}
& \plotzoomed{images/#1/normals.png}
& \plotall{images/#1/mapped_pano.png} \\ \vspace{-1cm}
\rotatebox{90}{~~~~~~~~~Ground Truth}
& \plotzoomed{images/#1/gt_final.png}
& \plotzoomed{images/#1/gt_normals.png}
& \plotall{images/#1/gt_pano.png} \\
& Novel View & Surface Normals & Environment Map
\end{tabular}
\captionof{figure}{Our method (top) recovers materials, geometry, and illumination that closely resemble the ground truth (bottom), optimizing directly from calibrated images of a scene. Here we show results on the \emph{#1} scene from NeRF~\cite{mildenhall2021nerf}: a rendered novel view (left), surface normals (middle), and environment map illumination (right). Insets show high-fidelity reflections, including interreflections, as well as accurate geometry even in concave regions. \todo{bottom of our surface normals are getting cut off}}
\label{fig:#1_teaser}
\end{center}
}]
}

\twocolumn[{
\renewcommand\twocolumn[1][]{#1}
\maketitle
\begin{center}
\includegraphics*[width=\linewidth]{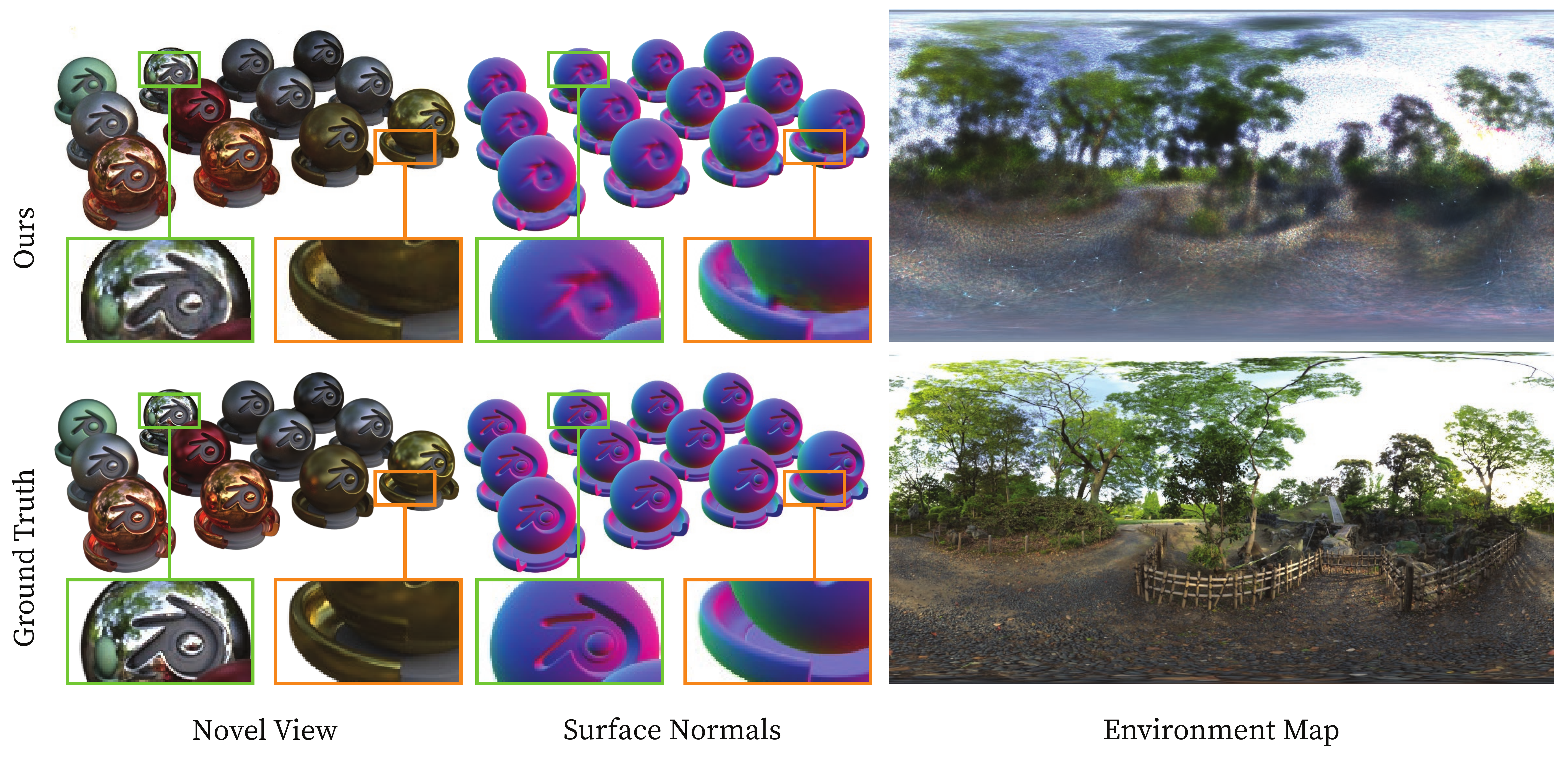}
\captionof{figure}{Our method (top) recovers materials, geometry, and illumination that closely resemble the ground truth (bottom), optimizing directly from calibrated images of a scene. Here we show results on the \emph{materials} scene from NeRF~\cite{mildenhall2021nerf}: a rendered novel view (left), surface normals (middle), and environment map illumination (right). Insets show high-fidelity reflections, including interreflections, as well as accurate geometry, even in concave regions.}
\label{fig:materials_teaser}
\end{center}
}]
\begin{abstract}

\vspace{-0.45cm}
\looseness=-1
We present \modelname{}, a method for recovering materials, geometry, and environment illumination from images of a scene. Our method uses a microfacet reflectance model within a volumetric setting by treating each sample along the ray as a (potentially non-opaque) surface.
Using surface-based Monte Carlo rendering in a volumetric setting enables our method to perform inverse rendering efficiently by combining decades of research in surface-based light transport with recent advances in volume rendering for view synthesis.
Our approach outperforms prior work in inverse rendering, capturing high fidelity geometry and high frequency illumination details; its novel view synthesis results are on par with state-of-the-art methods that do not recover illumination or materials.
\end{abstract}
\vspace{-0.2cm}

\section{Introduction}


Simultaneous recovery of the light sources illuminating a scene and the materials and geometry of objects inside it, given a collection of images, is a fundamental problem in computer vision and graphics. This decomposition enables editing and downstream usage of a scene: rendering it from novel viewpoints, and arbitrarily changing the scene's illumination, geometry, and material properties. This disentanglement is especially useful for creating 3D assets that can be inserted into other environments and realistically rendered under novel lighting conditions. 

Recent methods for novel view synthesis based on neural radiance fields~\cite{mildenhall2021nerf} have been highly successful at decomposing scenes into their geometry and appearance components, enabling rendering from new, unobserved viewpoints.
However, the geometry and appearance recovered are often of limited use in manipulating either materials or illumination, since they model each point as a direction-dependent emitter rather than as reflecting the incident illumination.
To tackle the task of further decomposing appearance into illumination and materials, we return to a physical model of light-material interaction, which models a surface as a distribution of microfacets that \emph{reflect} light rather than emitting it. By explicitly modeling this interaction during optimization, our method can recover both material properties and the scene's illumination.

\looseness=-1
Our method uses a Monte Carlo rendering approach with a hybrid surface-volume representation, where the scene is parameterized as a 3D field of microfacets: the scene's geometry is represented as a volume density, but its materials are parameterized using a spatially varying Bidirectional Reflectance Distribution Function (BRDF). 
The volumetric representation of geometry has been shown to be effective for optimization~\cite{mildenhall2021nerf,yariv2021volume}, and treating each point in space as a microfaceted surface allows us to use ideas stemming from decades of prior work on material parameterization and efficient surface-based rendering. 
Despite its volumetric parameterization, we verify experimentally that our model shrinks into a surface around opaque objects, with all contributions to the color of a ray coming from the vicinity of its intersection with the object.

To summarize, our method (1) combines aspects of volume-based and surface-based rendering for effective optimization, enabling reconstructing high-fidelity scene geometry, materials, and lighting from a set of calibrated images; (2) uses an optimizable microfacet material model rendered using Monte Carlo integration with multi-bounce raytracing, allowing for realistic interreflections on nonconvex objects; and (3) is efficient: it optimizes a scene from scratch in $\sim$3 hours on a single NVIDIA GTX 3090.

\section{Related work}

Our work lies in the rich field of inverse rendering, in which the goal is to reconstruct the geometry, material properties, and illumination that gave rise to a set of observed images. This task is a severely underconstrained inverse problem, with challenges ranging from lack of differentiability~\cite{tzumao} to the computational cost-variance tradeoff of the forward rendering process~\cite{hasselgren2022nvdiffrecmc}.

Recent progress in inverse rendering, and in particular in view synthesis, has been driven by modeling scenes as radiance fields \cite{mildenhall2021nerf, mipnerf, mipnerf360}, which can produce photorealistic models of a scene based on calibrated images.


\paragraph{Inverse rendering.}
Inverse rendering techniques can be categorized based on the combination of unknowns recovered and assumptions made. 
Common assumptions include far field illumination, isotropic BRDFs, and no interreflections or self-occlusions. Early work by Ramamoorthi and Hanrahan~\cite{ramamoorthi2001signal} handled unknown lighting, texture, and BRDF by using spherical harmonic representations of both BRDF and lighting, which allowed recovering materials and low frequency lighting components. More recent methods used differentiable rendering of known geometry, first through differentiable rasterization~\cite{loper2014opendr, Liu_2019_ICCV, chen2019learning} and later through differentiable ray tracing~\cite{tzumao, azinovic2019inverse, park2020seeing}. Later methods built on differentiable ray tracing, making use of Signed Distance Fields (SDFs) to also reconstruct geometry~\cite{zhang2021physg, munkberg2021nvdiffrec, hasselgren2022nvdiffrecmc}. 

Following the success of NeRF~\cite{mildenhall2021nerf}, volumetric rendering has emerged as a useful tool for inverse rendering.
Some methods based on volume rendering assume known lighting and only recover geometry and materials~\cite{srinivasan2021nerv, bi2020neural, asthana2022neural}, while others solve for both lighting and materials, but assume known geometry~\cite{lyu2022neural, zhang2021nerfactor} or geometry without self-occlusions~\cite{boss2021neural, boss2021nerd}. Some methods simultaneously recover illumination, geometry, and materials, but assume that illumination comes from a single point light source~\cite{guo2020object,iron-2022}. However, to the best of our knowledge none of these existing volumetric inverse rendering methods are able to capture high frequency lighting (and appearance of specular objects) from just the input images themselves. 

An additional challenge is in handling multi-bounce illumination, or interreflections, in which light from a source bounces off multiple objects before reaching the camera. In this case, computational tradeoffs are unavoidable due to the exponential growth of rays with the number of such bounces. Park \etal~\cite{park2020seeing} model interreflections assuming known geometry, but do not model materials, which is equivalent to treating all objects as perfect mirrors. Other methods use neural networks to cache visibility fields~\cite{srinivasan2021nerv, zhang2021nerfactor} or radiance transfer fields~\cite{lyu2022neural, guo2020object}. Our method handles interreflections by casting additional rays through the scene, using efficient Monte Carlo sampling.

\paragraph{Volumetric view synthesis.}
We base our representation of geometry on recent advances in volumetric view synthesis, following NeRF~\cite{mildenhall2021nerf}. Specifically, we retain the idea of using differentiable volumetric rendering to model geometry, using a voxel-based representation of the underlying density field~\cite{plenoxels, ingp, tensorf}.

Of particular relevance to our work are prior methods such as~\cite{verbin2021ref,ge2023refneus} that are specifically designed for high-fidelity appearance of glossy objects. In general, most radiance field models fail at rendering high-frequency appearance caused by reflections from shiny materials under natural illumination, instead rendering blurry appearance~\cite{zhang2020nerf++}. To enable our method to handle these highly specular materials, and to improve the normal vectors estimated by our method, which are key to the rendered appearance, we utilize regularizers from Ref-NeRF~\cite{verbin2021ref}.

\section{Preliminaries}

Our method combines aspects of \emph{volumetric} and \emph{surface-based} rendering; we begin with a brief introduction to each before describing our method in Section~\ref{sec:method}.

\subsection{Volume Rendering}

The core idea in emission-absorption volume rendering is that light accumulates along rays, with ``particles'' along the ray both emitting and absorbing light. The color measured by a camera pixel corresponding to a ray with origin $\point_c$ and direction $-\omegao$ is:
\begin{align} \label{eqn:volrendering2}
    L(\point_c, \omegao) &= \int_0^\infty T(t) \density(\ray(t))L_o(\ray(t), \omegao) dt, \\
    \text{where}\;\; T(t) &= \exp\left( -\int_0^{t}\density(\ray(t'))dt' \right),    
\end{align}
where $\ray(t)=\point_c - t\omegao$ is a camera ray, $\density(\point)$ is the density at point $\point$ in the volume, $T$ denotes transmittance along the ray, and $L_o$ is the outgoing radiance. This formula is often approximated numerically using quadrature, following~\cite{max1995}:
\begin{align}
L(\point_c, \omegao) &\approx \sum_{j=0}^{N-1} w_j L_o(\ray(t_j), \omegao), \\
\label{eqn:volrenderingquadrature}
\text{where}\;\; w_j &= T_j \big(1-\exp(-\sigma(\ray(t_j))(t_{j+1}-t_j))\big), \\
\text{and}\;\; T_j &= \exp\left(-\sum_{k=0}^{j-1}\sigma(\ray(t_k))(t_{k+1}-t_k) \right).
\end{align}
In this volume rendering paradigm, multiple 3D points can contribute to the color of a ray, with nearer and denser points contributing most.

\subsection{Surface Rendering}

In surface rendering, and assuming fully-opaque surfaces, the color of a ray is determined solely by the light reflected by the first surface it encounters. Consider that the ray $\ray$ from camera position $\point_c$ in direction $-\omegao$ intersects its first surface at a 3D position $\point$. The ray color is then: 
\begin{equation}
    \label{eqn:rendering}
    L(\point_c, \omegao) = \int_{\mathbb{S}^2} f(\point, \omegao, \omegai)L_i(\point, \omegai) (\uvn(\point)\cdot\omegai)^+ d\omegai,
\end{equation}
where $\omegai$ is the direction of incident light, $\uvn(\point)$ is the surface normal at $\point$, $f$ is the \bsdf{} describing the material of the surface at $\point$, $L_i$ is the incident radiance, and $(\uvn(\point)\cdot\omegai)^+$ is a truncated cosine lobe (\ie its negative values are clipped to zero) facing outward from the surface. Note that this equation is recursive: $L_i$ inside the integral may be the outgoing radiance $L_o$ coming from a different scene point.
The integral in Equation~\ref{eqn:rendering} is also typically approximated by discrete (and often random) sampling, and is the subject of a rich body of work \cite{cook1982reflectance, veach1998robust}.


\begin{figure*}[ht]
    \centering
    \includegraphics[width=\textwidth]{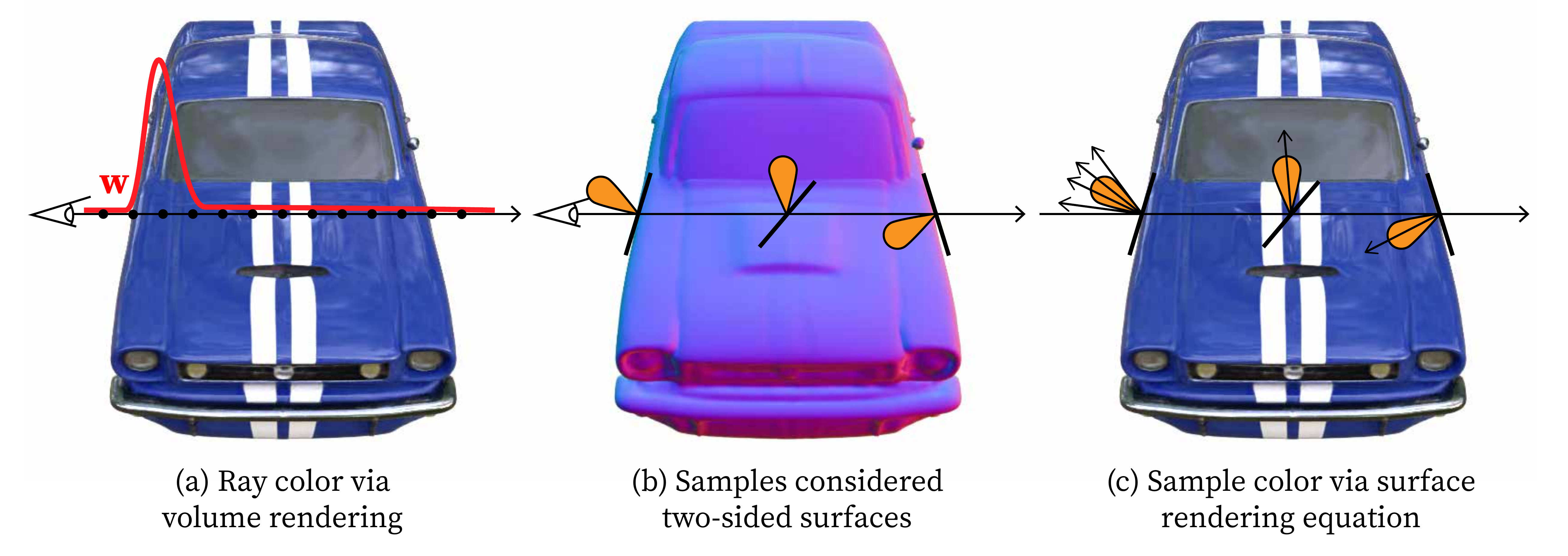}
    \caption{To render the color of a ray cast through the scene, we (a) evaluate density at each sample and compute each sample's volume rendering quadrature weight $w_i$, then (b) query the material properties and surface normal (flipped if it does not face the camera) at each sample point, which are used to (c) compute the color of each sample by using Monte Carlo integration of the surface rendering integral, where the number of samples used is proportional to the quadrature weight $w_i$. This sample color is then accumulated along the ray using the quadrature weight to get the final ray color.
    }
    \label{fig:methoddiagram}
\end{figure*}

\section{Method} \label{sec:method}

We present \modelname{} to tackle the problem of inverse rendering by combining volume and surface rendering, as shown in Figure~\ref{fig:methoddiagram}. Our method takes as input a collection of images ($100$ in our experiments) with known cameras, and outputs the volumetric density and normals, materials (\bsdf{}s), and far-field illumination (environment map) of the scene. We assume that all light sources are infinitely far away from the scene, though light may interact locally with multiple bounces through the scene.

\looseness=-1
In this section, we describe our representation of a scene and the rendering pipeline we use to map this representation into pixel values. 
Section~\ref{sec:main idea} introduces the main idea of our method, to build intuition before diving into the details. 
Section~\ref{section:geometry} describes our representation of the scene geometry, including density and normal vectors. Section~\ref{section:BRDF} describes our representation of materials and how they reflect light via the \bsdf{}.
Section~\ref{section:illumination} introduces our parameterization of illumination, which is based on a far-field environment map equipped with an efficient integrator for faster evaluations of the rendering integral. Finally, Section~\ref{section:rendering} describes the way we combine these different components to render pixel values in the scene.

\subsection{Main Idea} \label{sec:main idea}

\begin{figure*}[ht]
    \centering
    \includegraphics[width=\textwidth]{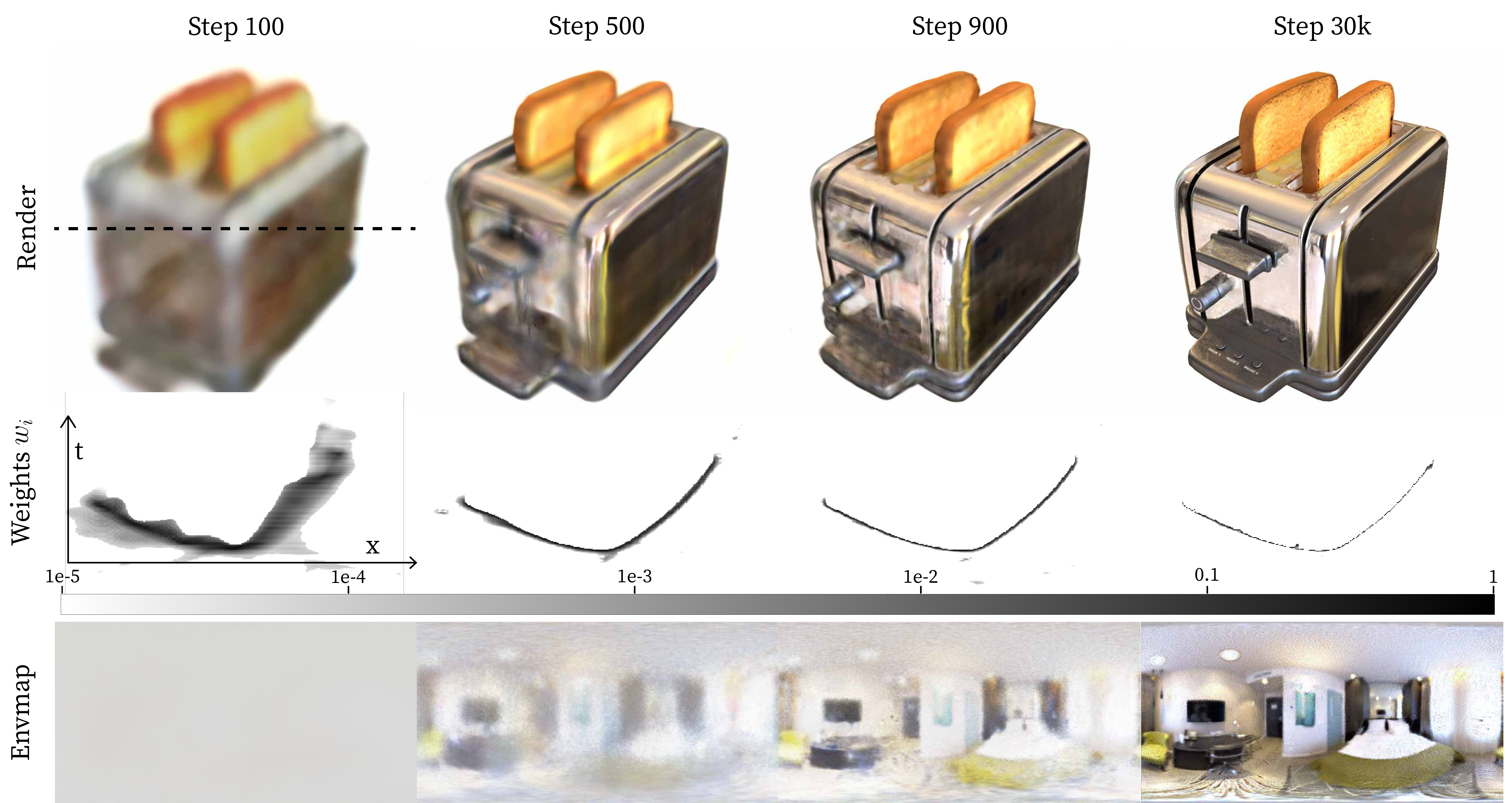}
    \caption{\textbf{Snapshots of the toaster scene during optimization.} The second row shows a cross section of the weights along each ray taken along the dotted line. Early in training the object geometry is cloudy and the environment map is uniform, but as training proceeds the object develops a sharp surface and the environment map converges.}
    \label{fig:toaster over time}
\end{figure*}

The key to our method is a novel combination of the volume rendering and surface rendering paradigms: we model a density field as in volume rendering, and we model outgoing radiance at every point in space using surface-based light transport (approximated using Monte Carlo ray sampling).
Volume rendering with a density field lends itself well to optimization: initializing geometry as a semi-transparent cloud creates useful gradients (see Figure~\ref{fig:toaster over time}), and allows for changes in geometry and topology. Using surface-based rendering allows modeling the interaction of light and materials, and enables recovering these materials.

We combine these paradigms by modeling a \emph{microfacet field}, in which each point in space is endowed with a volume density and a local micro-surface. Light accumulates along rays according to the volume rendering integral of Equation~\ref{eqn:volrendering2}, but the outgoing light of each 3D point is determined by surface rendering as in Equation~\ref{eqn:rendering}, using rays sampled according to its local micro-surface. 
This combination of volume-based and surface-based representation and rendering, shown in Figure~\ref{fig:methoddiagram}, enables us to optimize through a severely underconstrained inverse problem, recovering geometry, materials, and illumination simultaneously. 

\subsection{Geometry Parameterization} \label{section:geometry}

We represent geometry using a low-rank tensor data structure based on TensoRF~\cite{tensorf}, with small modifications described in 
Appendix~\ref{appendix:optimization}. 
Our model stores both density $\density$ and a spatially-varying feature that is decoded into the material's \bsdf{} at every point in space. We initialize our model at low resolution and gradually upsample it during optimization (see Appendix~\ref{appendix:optimization} for details).

Similar to prior work~\cite{srinivasan2021nerv,verbin2021ref}, we use the negative normalized gradient of the density field as a field of ``volumetric normals.''
However, like~\cite{kuang2022neroic}, we found that numerically computing spatial gradients of the density field using finite differences rather than using analytic gradients leads to normal vectors that we can use directly, without using features predicted by a separate MLP. Additionally, these numerical gradients can be efficiently computed using 2D and 1D convolution using TensorRF's low-rank density decomposition (see Appendix~\ref{appendix:optimization}). These accurate normals are then used for rendering the appearance at a volumetric microfacet, as will be discussed in Sections~\ref{section:BRDF} and~\ref{section:rendering}. 


Our volumetric normals are regularized using the orientation loss $\mathcal{R}_o$ introduced by Ref-NeRF~\cite{verbin2021ref}:
\begin{equation}
    \label{eqn:normal_penalty}
    \mathcal{R}_o = \sum_j w_j \max(0, -\uvn(\point_j) \cdot \omegao)^2,
\end{equation}
where $\omegao$ is the view direction facing towards the camera, and $\uvn(\point_j)$ is the normal vector at the $j$th point along the ray. The orientation loss $\mathcal{R}_o$ penalizes normals that face away from the camera yet contribute to the color of the ray (as quantified by weights $w_j$).

Because our volumetric normals are derived from the density field, this regularizer has a direct effect on the reconstructed geometry: it decreases the weight of backwards-facing normals by decreasing their density or increasing the density between them and the cameras, thereby promoting hard surfaces and improving reconstruction. 
Note that unlike Ref-NeRF, we do not use ``predicted normals'' for surface rendering, as our Gaussian-smoothed derivative filter achieves similar effect.

\subsection{Material Representation}  \label{section:BRDF}

We write our spatially varying \bsdf{} model $f$ 
as a combination of diffuse and specular components:
\begin{align}
    f(\omegao, \omegai) =
    \frac{\rho}{\pi}(1-F_r(\hb)) + F_r(\hb) f_s&(\omegao, \omegai),
\end{align}
where $\rho$ is the RGB albedo,
$\hb=\frac{\omegao+\omegai}{\|\omegao+\omegai\|}$ is the half vector,
$F_r(\hb)$ is the Fresnel term, $f_s(\omegao, \omegai)$ is the specular component of the \bsdf{}
for outgoing view direction $\omegao$ and incident light direction $\omegai$. The spatial dependence of these terms on the point $\point$ is omitted for brevity. We use the Schlick approximation~\cite{schlick1994inexpensive} for the Fresnel term:
\begin{equation}
    \label{eqn:fresnel}
    F_r(\hb) = F_0(\point) + (1-F_0(\point))(1-\hb\cdot \omegao)^5,
\end{equation}
where $F_0(\point) \in [0, 1]^3$ is the spatially varying reflectance at the normal incidence at the point $\point$, and we base our specular \bsdf{} on the Cook-Torrance \bsdf~\cite{torrance1967theory}, using:
\begin{equation}
    \label{eqn:brdf}
    f_s(\omegao, \omegai) = \frac{D(\hb ; \alpha, \uvn, \omegao)G_1(\omegao, \hb)g(\omegai, \omegao)}{4\left(\uvn\cdot\omegao\right)\left(\uvn\cdot\omegai\right)
    },
\end{equation}
where $D$ is a Trowbridge-Reitz distribution~\cite{trowbridge1975average} (popularized by the GGX \bsdf{} model~\cite{walter2007microfacet}), $G_1$ is the Smith shadow masking function for the Trowbridge-Reitz distribution, and $g$ is a shallow multilayer perceptron (MLP) with a sigmoid nonlinearity at its output. 
The distribution $D$ models the roughness $\alpha$ of the material, 
and it is used for importance sampling, as described in Section~\ref{section:rendering}). The MLP $g$ captures other material properties not included in its explicit components.

The parameters for each of the diffuse and specular \bsdf{} components are stored as features $\feat$ in the TensoRF representation (alongside density $\density$), allowing them to vary in space. We compute the roughness $\alpha$, albedo $\rho$ and reflectance at the normal incidence $F_0$ by applying a single linear layer with sigmoid activation to the spatially-localized features $\feat$. 
Details about the architecture of the MLP $g$ and its input encoding can be found in Appendix~\ref{appendix:parameterization}.

We can approximately evaluate the rendering equation integral in Equation~\ref{eqn:rendering} more efficiently by assuming all microfacets at a point $\point$ have the same irradiance $E(\point)$:
\begin{align}
    \int_{\mathbb{S}^2} &f(\point, \omegao, \omegai)L_i(\point, \omegai) (\uvn(\point)\cdot\omegai)^+ d\omegai  \\
    \label{eqn:approx_render}
    \approx \int_{\mathbb{S}^2} &F_r(\hb)f_s(\point, \omegao, \omegai)L_i(\point, \omegai) (\uvn(\point)\cdot\omegai)^+ d\omegai  \nonumber \\
    & + \frac{\rho(\point)}{\pi}E(\point) \int_{\mathbb{S}^2}(1-F_r(\hb))d\omegai .
\end{align}
The irradiance $E(\point)$, defined as:
\begin{equation}
    E(\point) = \int_{\mathbb{S}^2} L_i(\point, \omegai) (\uvn(\point)\cdot\omegai)^+ d\omegai ,
\end{equation}
can then be easily evaluated using an irradiance environment map approximated by low-degree spherical harmonics, as done by Ramamoorthi and Hanrahan~\cite{ramamoorthi2001efficient}. At every optimization step, we obtain the current irradiance environment map by integrating the environment map with spherical harmonic functions up to degree $2$, and combining the result with the coefficients of a clamped cosine lobe pointing in direction $\uvn(\point)$ to obtain the irradiance $E(\point)$~\cite{ramamoorthi2001efficient}.
Equation~\ref{eqn:approx_render} can then be importance sampled according to $D$ and integrated using Monte Carlo sampling of incoming light. 

We sample half vectors $\hb$ from the distribution of visible normals $D_{\omegao}$~\cite{heitz2018sampling}, which is defined as:
\begin{equation} \label{eqn:visible}
    D_{\omegao}(\hb) = \frac{G_1(\omegao, \hb)|\omegao\cdot\hb|D(\hb)}{|\omegao\cdot\uvn|} .
\end{equation}
However, to perform Monte Carlo integration of the rendering equation, we need to convert from half vector space to the space of incoming light, which requires multiplying by the determinant of the Jacobian of the reflection equation $\omegai = 2(\omegao\cdot \hb) \hb - \omegao$, which is $4 (\omegao\cdot\hb)$~\cite{walter2007microfacet}. Multiplying by the $\omegai\cdot\uvn$ 
term from Equation~\ref{eqn:rendering} as well as the Jacobian and Equation~\ref{eqn:visible} results in the following Monte Carlo estimate:
\begin{align}
    \label{eqn:montecarlo}
    L(\point, \omegao) \approx 
    \frac 1{N}\sum_{n=1}^{N} &F_r(\hat{\mathbf{h}}^n)g(\feat, \omegao, \omegan)L_i(\point, \omegan)\nonumber  \\
    & + (1-F_r(\hat{\mathbf{h}}^n)) \frac{\rho(\point)}{\pi}E(\point), \\
    \text{where }\hat{\mathbf{h}}^n \sim &D_{\omegao}(\,\,\cdot\,\,; \alpha(\point), \uvn(\point), \omegao), \nonumber \\
    \text{and }\omegan = &2(\omegao\cdot \hat{\mathbf{h}}^n) \hat{\mathbf{h}}^n - \omegao. \nonumber
\end{align}

\subsection{Illumination} \label{section:illumination}

We model far field illumination using an environment map, represented using an equirectangular image with dimensions $H\times W\times 3$, with $H=512$ and $W=1024$.
\looseness=-1
We map the optimizable parameters in our environment map parameterization into high dynamic range RGB values by applying an elementwise exponential function.

We use the term \emph{primary} to denote a ray originating at the camera, and \emph{secondary} to denote a ray bounced from a surface to evaluate its reflected light (whether that light arrives directly from the environment map, or from another scene point).

To minimize sampling noise, instead of using a single environment map element per secondary ray, we use the mean value over an axis-aligned rectangle in spherical coordinates, with the solid angle covered by the rectangle adjusted to match the sampling distribution $D$ at that point.  Concretely, to query the environment map at a given incident light direction, we first compute its corresponding spherical coordinates $(\theta, \phi)$, where $\theta$ and $\phi$ are the polar and azimuthal angles respectively. We then compute the mean value of the environment map over a (spherical) rectangle centered at $(\theta, \phi)$, whose size $\Delta\theta\times\Delta\phi$ we constrain to have aspect ratio $\frac{\Delta\theta}{\Delta\phi} = \sin\theta$. To choose the solid angle of the rectangle, $\Delta\theta\cdot\Delta\phi$, we modify the method from~\cite{colbert2007gpu}, which is based on Nyquist's sampling theorem (see Appendix~\ref{appendix:nyquist}). 
We compute these mean values efficiently using integral images, also known as summed-area tables~\cite{crow1984summed}.



\subsection{Rendering} \label{section:rendering}

\looseness=-1
In this section, we describe how the model components representing geometry (Section~\ref{section:geometry}), materials (Section~\ref{section:BRDF}), and illumination (Section~\ref{section:illumination}) are combined to render the color of a pixel. 
For each primary ray, we choose a set of sample points following the rejection-sampling strategy of~\cite{li2022nerfacc, ingp} to prioritize points near object surfaces. We query our geometry representation~\cite{tensorf} for the density $\sigma_j$ at each point, and use Equation~\ref{eqn:volrenderingquadrature} to estimate the contribution weight $w_j$ of each point to the final ray color.


To compute the color for each of these points, we compute the irradiance from the environment map and apply Equation~\ref{eqn:montecarlo} to obtain $L(\point_j, \omegao)$, as described in Section~\ref{section:BRDF}.

For each primary ray sample with weight $w_j$, we allocate $N=\floor*{w_j M}$ secondary rays, where $M=128$ is an upper bound on the total number of secondary rays for each primary ray (since $\sum_j w_j \leq 1$). The secondary rays are sampled according to the Trowbridge-Reitz distribution
using the normal vector $\uvn(\point_j)$ and roughness value $\alpha_j$ at the current sample.

When computing the incoming light $L_i(\point, \omegan)$, we save memory by randomly selecting a fixed number $R$ of secondary rays to interreflect through the scene while others index straight into the environment map.
We importance sample these $R$ secondary rays according to the largest channel of the weighted RGB multiplier $w_j\cdot g(\feat_j, \omegai, \omegao)$. In practice, we find that adding a small amount of random noise to the weighted RGB multiplier before choosing the $R$ largest values improves performance by slightly increasing the variation of the selected rays. 

The remaining $N-R$ secondary rays with lower contribution are rendered more cheaply by evaluating the environment map directly rather than considering further interactions with the scene, as described in Section~\ref{section:illumination}.

This combined sample color value $L(\point_j, \omegao)$ is then weighted by $w_j$, and the resulting colors are summed along the primary ray samples to produce pixel values.
Finally, the resulting pixel values are tonemapped to sRGB color and clipped to $[0, 1]$.

\paragraph{Dynamic batching.}
We apply the dynamic batch size strategy from NeRFAcc~\cite{li2022nerfacc} to the TensoRF~\cite{tensorf} sampler, which controls the number of samples per batch using the number of primary rays per batch. Since the number of secondary rays scales with the number of primary rays, we bound the maximal number of primary rays to avoid casting too many secondary rays. We use the same method to control $R$, the number of secondary bounces that retrace through the scene. During test time, we shuffle the image to match the training distribution, then unshuffle the image to get the result.

We train using photometric loss and the normal penalty loss of Equation~\ref{eqn:normal_penalty}. 
Further details of our optimization and sampling methods can be found in Appendix~\ref{appendix:optimization}.

\section{Experiments}

We evaluate our method using two synthetic datasets: the \emph{Blender} dataset from NeRF~\cite{mildenhall2021nerf}, and the \emph{Shiny Blender} dataset from Ref-NeRF~\cite{verbin2021ref}. Both datasets contain objects rendered against a white background; the \emph{Shiny Blender} dataset focuses on shiny materials with high-frequency reflections whereas the \emph{Blender} dataset contains a mixture of specular, glossy, and Lambertian materials. 

We evaluate standard metrics PSNR, SSIM \cite{wang2004image}, and LPIPS \cite{zhang2018unreasonable} on the novel view synthesis task for each dataset. To quantify the quality of our reconstructed geometry, we also evaluate the Mean Angular Error (MAE$^\circ$) of the normal vectors. MAE$^\circ$ is evaluated over the same test set of novel views; for each view we take the dot product between the ground truth and predicted normals, multiply the $\arccos$ of this angle by the ground truth opacity of the pixel, and take the mean over all pixels. This gives us an error of $90^\circ$ if the predicted normals are missing (\ie if the pixel is mistakenly predicted as transparent). We summarize these quantitative metrics in Table~\ref{tab:combined}, with per-scene results in Appendix~\ref{appendix:results}.

We provide qualitative comparisons of our reconstructed environment maps with prior inverse rendering approaches in Figures~\ref{fig:car} and~\ref{fig:helmet}, and additional results in Appendix~\ref{appendix:results}. We also demonstrate two applications of our scene decomposition in Figure~\ref{fig:environment swap}. For Figure~\ref{fig:environment swap} (a), we render the geometry and spatially varying BRDF recovered from the \emph{materials} scene with the environment map optimized from the \emph{helmet} scene, showing convincing new reflections while retaining the original object shape and material properties. For Figure~\ref{fig:environment swap} (b), we take this a step further by training the \emph{toaster} and \emph{car} scenes with the same neural material decoder, which enabled us to compose the two scenes under the environment map recovered from the \emph{toaster} scene.

To quantitatively measure the quality of our method's disentanglement, we evaluate it similarly to NeRFactor~\cite{zhang2021nerfactor}. Since our method is specifically designed to handle more specular objects, we modify the Shiny Blender dataset from Ref-NeRF~\cite{verbin2021ref} by rendering the images in HDR under both the original lighting and an unseen lighting condition. We then optimize each method on each of the scenes with original lighting, then render the estimated geometry and materials using the unseen lighting condition. For computing quality metrics, we find the (ambiguous) per-channel scaling factor by minimizing the mean squared error. The scaled image is then evaluated against the ground truth relit image using PSNR, SSIM, and LPIPS. 

The results are presented in Table~\ref{tab:relighting}, showing that our method is able to relight shiny objects with significantly increased accuracy when compared to state of the art inverse rendering methods such as NVDiffRec and NVDiffRecMC.

\plotenvswap{}

Table~\ref{tab:combined} also shows quantitative ablation studies on our model. If we do not use integral images for the far field illumination (``no integral image''), the sparse gradient on the environment map prevents the model from learning to use it to explain reflections.
Using the derivative of the linear interpolation of the density instead of smoothed numerical derivatives (``analytical grad''), results in more holes in the geometry, limiting performance.
If we do not utilize multiple ray bounces for interreflections (``single bounce''), all metrics are slightly worse, with most of the errors arising in regions with strong interreflections.
Finally, replacing the neural network with the identity function (``no neural''), results in slightly worse performance, especially in regions with strong interreflections.



\plotsceneshort{car} 

\plotscene{helmet}

\begin{table*}[]
\centering
\resizebox{0.8\linewidth}{!}{
\begin{tabular}{l|rrrr|rrrr}
                            & \multicolumn{1}{l}{Blender}         & \multicolumn{1}{l}{}                & \multicolumn{1}{l}{}                   & \multicolumn{1}{l|}{}                              & \multicolumn{1}{l}{Shiny Blender}   & \multicolumn{1}{l}{}                & \multicolumn{1}{l}{}                   & \multicolumn{1}{l}{}                              \\ \hline
                            & \multicolumn{1}{l}{PSNR $\uparrow$} & \multicolumn{1}{l}{SSIM $\uparrow$} & \multicolumn{1}{l}{LPIPS $\downarrow$} & \multicolumn{1}{l|}{$\text{MAE}^\circ \downarrow$} & \multicolumn{1}{l}{PSNR $\uparrow$} & \multicolumn{1}{l}{SSIM $\uparrow$} & \multicolumn{1}{l}{LPIPS $\downarrow$} & \multicolumn{1}{l}{$\text{MAE}^\circ \downarrow$} \\ \hline
PhySG$^1$                   & 18.54                               & .847                                & .182                                   & 29.17                                              & 26.21                               & .921                                & .121                                   & 8.46                                              \\
NVDiffRec$^1$               & 28.79                               & .939                                & .068                                   & 11.788                                             & 29.90                               & .945                                & .114                                   & 31.885                                            \\
NVDiffRecMC$^1$             & 25.81                               & .904                                & .111                                   & 9.003                                              & 28.20                               & .902                                & .175                                   & 28.682                                            \\
Ref-NeRF$^2$                & \cellcolor[HTML]{FFB3B3}33.99       & \cellcolor[HTML]{FFB3B3}.966        & \cellcolor[HTML]{FFB3B3}.038           & 23.22                                              & \cellcolor[HTML]{FFB3B3}35.96       & \cellcolor[HTML]{FFB3B3}.967        & \cellcolor[HTML]{FFD9B3}.059           & 18.38                                             \\
Ours, no integral image     & 28.47                               & .920                                & .069                                   & 27.988                                             & 27.33                               & .869                                & .170                                   & 34.118                                            \\
Ours, analytical derivative & 28.94                               & .926                                & .064                                   & 13.976                                             & 29.22                               & .900                                & .140                                   & 24.295                                            \\
Ours, single bounce         & \cellcolor[HTML]{FFFFB4}30.68       & \cellcolor[HTML]{FFD9B3}.944        & \cellcolor[HTML]{FFD9B3}.045           & \cellcolor[HTML]{FFD9B3}6.216                      & \cellcolor[HTML]{FFFFB4}34.39       & \cellcolor[HTML]{FFFFB4}.962        & \cellcolor[HTML]{FFB3B3}.053           & \cellcolor[HTML]{FFD9B3}17.647                    \\
Ours, no neural             & 29.40                               & .933                                & .057                                   & \cellcolor[HTML]{FFFFB4}7.325                      & 33.00                               & .955                                & \cellcolor[HTML]{FFFFB4}.063           & \cellcolor[HTML]{FFFFB4}19.190                    \\
Ours                        & \cellcolor[HTML]{FFD9B3}30.71       & \cellcolor[HTML]{FFFFB4}.940        & \cellcolor[HTML]{FFFFB4}.053           & \cellcolor[HTML]{FFB3B3}6.061                      & \cellcolor[HTML]{FFD9B3}34.56       & \cellcolor[HTML]{FFD9B3}.963        & \cellcolor[HTML]{FFB3B3}.053           & \cellcolor[HTML]{FFB3B3}17.497                   
\end{tabular}
}\\
\small{$^1$ requires object masks during training. ~~ $^2$ view synthesis method, not inverse rendering. ~~Red is best, followed by orange, then yellow. }
\caption{\textbf{Results on the \emph{Blender} dataset from NeRF \cite{mildenhall2021nerf}. and \emph{Shiny Blender} dataset from Ref-NeRF \cite{verbin2021ref}.} We compute PSNR, SSIM, and LPIPS on the novel view synthesis task, and MAE$^\circ$ on the normals. Our method outperforms all prior methods except Ref-NeRF on view synthesis, and produces the most accurate geometric normals.}
\label{tab:combined}
\end{table*}

\begin{table}[t]
\resizebox{\linewidth}{!}{
\begin{tabular}{l|rrrrrrr}
PSNR $\uparrow$    & \multicolumn{1}{l}{toaster}   & \multicolumn{1}{l}{coffee}    & \multicolumn{1}{l}{helmet}    & \multicolumn{1}{l}{ball}      & \multicolumn{1}{l}{teapot}    & \multicolumn{1}{l}{car}       & \multicolumn{1}{l}{mean}      \\ \hline
NVDiffRec$^1$      & \cellcolor[HTML]{FFFFB4}12.56 & 20.30                         & \cellcolor[HTML]{FFFFB4}17.73 & 13.21                         & \cellcolor[HTML]{FFFFB4}33.00 & 19.84                         & 19.44                         \\
NVDiffRecMC$^1$    & \cellcolor[HTML]{FFD9B3}14.55 & \cellcolor[HTML]{FFD9B3}23.39 & \cellcolor[HTML]{FFD9B3}20.44 & \cellcolor[HTML]{FFFFB4}13.23 & 32.51                         & \cellcolor[HTML]{FFFFB4}20.09 & \cellcolor[HTML]{FFFFB4}20.70 \\
Ours               & \cellcolor[HTML]{FFB3B3}20.05 & \cellcolor[HTML]{FFFFB4}24.21 & \cellcolor[HTML]{FFB3B3}25.69 & \cellcolor[HTML]{FFB3B3}23.50 & \cellcolor[HTML]{FFD9B3}34.37 & \cellcolor[HTML]{FFB3B3}23.19 & \cellcolor[HTML]{FFB3B3}25.17 \\ \hline
LPIPS $\downarrow$ & \multicolumn{1}{l}{toaster}   & \multicolumn{1}{l}{coffee}    & \multicolumn{1}{l}{helmet}    & \multicolumn{1}{l}{ball}      & \multicolumn{1}{l}{teapot}    & \multicolumn{1}{l}{car}       & \multicolumn{1}{l}{mean}      \\ \hline
NVDiffRec$^1$      & .378                          & .208                          & .229                          & .467                          & .027                          & .113                          & .237                          \\
NVDiffRecMC$^1$    & \cellcolor[HTML]{FFFFB4}.290  & \cellcolor[HTML]{FFFFB4}.177  & \cellcolor[HTML]{FFFFB4}.217  & \cellcolor[HTML]{FFFFB4}.423  & \cellcolor[HTML]{FFFFB4}.024  & \cellcolor[HTML]{FFFFB4}.105  & \cellcolor[HTML]{FFFFB4}.206  \\
Ours               & \cellcolor[HTML]{FFB3B3}.170  & \cellcolor[HTML]{FFD9B3}.142  & \cellcolor[HTML]{FFB3B3}.129  & \cellcolor[HTML]{FFB3B3}.191  & \cellcolor[HTML]{FFB3B3}.017  & \cellcolor[HTML]{FFB3B3}.063  & \cellcolor[HTML]{FFB3B3}.119  \\ \hline
SSIM $\uparrow$    & \multicolumn{1}{l}{toaster}   & \multicolumn{1}{l}{coffee}    & \multicolumn{1}{l}{helmet}    & \multicolumn{1}{l}{ball}      & \multicolumn{1}{l}{teapot}    & \multicolumn{1}{l}{car}       & \multicolumn{1}{l}{mean}      \\ \hline
NVDiffRec$^1$      & .624                          & .870                          & .824                          & .626                          & .985                          & .853                          & .797                          \\
NVDiffRecMC$^1$    & \cellcolor[HTML]{FFD9B3}.741  & \cellcolor[HTML]{FFFFB4}.913  & \cellcolor[HTML]{FFFFB4}.861  & \cellcolor[HTML]{FFFFB4}.761  & \cellcolor[HTML]{FFFFB4}.986  & \cellcolor[HTML]{FFFFB4}.869  & \cellcolor[HTML]{FFFFB4}.855  \\
Ours               & \cellcolor[HTML]{FFB3B3}.864  & \cellcolor[HTML]{FFB3B3}.910  & \cellcolor[HTML]{FFB3B3}.918  & \cellcolor[HTML]{FFB3B3}.897  & \cellcolor[HTML]{FFD9B3}.987  & \cellcolor[HTML]{FFB3B3}.917  & \cellcolor[HTML]{FFB3B3}.916 
\end{tabular}
}
\small{$^1$ requires object masks during training. ~~Red is best, followed by orange, then yellow. }
\caption{\textbf{Relighting on the \emph{Shiny Blender} dataset from Ref-NeRF~[35].} 
}
\label{tab:relighting}
\end{table}







\section{Discussion and Limitations}

We introduced a novel and successful approach for the task of inverse rendering, using calibrated images alone to decompose a scene into its geometry, far-field illumination, and material properties. Our approach uses a combination of volumetric and surface-based rendering, in which we endow each point in space with both a density and a local microsurface, so that it can both occlude and reflect light from its environment. 
We verified experimentally that our method, which enjoys both the optimization landscape of volume rendering, as well as the richness and efficiency of surface-based Monte Carlo rendering, provides superior results relative to prior work. 

However, our method is not without limitations. 
First, although it can handle non-convex geometry, it assumes far field illumination and thus performs poorly when this assumption is not satisfied. This issue is most clear in the \emph{coffee} scene in the Shiny Blender dataset, which has near-field light sources. It also does not handle interreflections very well, since the number of secondary bounces is limited, and due to our acceleration scheme of often directly querying the environment map, as explained in Section~\ref{section:rendering}.
Our model also does not handle refractive media,
which is most clear in the \emph{drums} and \emph{ship} scenes in the Blender dataset.
Our diffuse lighting model also assumes far field illumination, and thus fails to fully isolate shadows from the albedo, most obvious in the \emph{lego} scene in the Blender dataset. These scenes are visualized in Appendix~\ref{appendix:results}.
Another limitation of our model's \bsdf{} parameterization is that it struggles to represent anisotropic materials. 
Finally, our method exhibits some speckle noise in its renderings, particularly near bright spots, which may be alleviated by using a denoiser as was used in~\cite{hasselgren2022nvdiffrecmc}.
We believe these limitations would make for interesting future work, as well as applying our method to other field representations and larger scenes captured in the wild.


\section{Acknowledgements}
We would like to thank Tzu-Mao Li for his advice, especially regarding the BRDF formulation, as well as his La Jolla renderer, which we used as a reference. AM is supported by ALERTCalifornia, developing technology to stay ahead of disasters, and the National Science Foundation under award \#CNS-1338192, MRI: Development of Advanced Visualization Instrumentation for the Collaborative Exploration of Big Data, and Kinsella Expedition Fund. This material is partially based upon work supported by the National Science Foundation under Award No. 2303178 to SFK.


{\small
\bibliographystyle{ieee_fullname}
\bibliography{egbib}
}

\clearpage

\appendix
\section{Rectangle Size Derivation} \label{appendix:nyquist}

As mentioned in Section~\ref{section:illumination} of the main paper, a ray that reaches the environment map is assigned a color taken as the average color over an axis-aligned rectangle in spherical coordinates, where the shape of the rectangle depends on the ray's direction and the material's roughness at the ray's origin. 
We modify the derivation of the area of the rectangle from GPU Gems~\cite{colbert2007gpu}. Let $N$ be the number of samples, $p(\omegai)$ be the probability density function of a given sample direction $\omegai$ for viewing direction $\omegao$, and let $H$ and $W$ be the height and width of the environment map (\ie its polar and azimuthal resolutions). The density $d(\omegai)$ of environment map pixels at a given direction must be inversely proportional to the Jacobian's determinant, $\sin\theta_i$, and it must also satisfy: 
\begin{equation}
    HW = \int_0^{2\pi}\int_0^\pi d(\omegai) \sin\theta_i d\theta_i d\phi_i,
\end{equation}
and therefore:
\begin{equation}
    d(\omegai) = \frac{HW}{2\pi^2\sin\theta_i}.
\end{equation}

The number of pixels per sample, which is the area of the rectangle, is then the total solid angle per sample, $Np(\omegai)$ multiplied by the number of pixels per solid angle:
\begin{equation}
    \Delta\theta\cdot \Delta\phi = \frac{Np(\omegai)}{d(\omegai)},
\end{equation}
where $\Delta\theta$ is the polar size of the rectangle, and $\Delta\phi$ is its azimuthal size, \ie the rectangle is $\Delta\theta\times\Delta\phi$, in equirectangular coordinates.

As mentioned in Section~\ref{section:illumination} of the main paper, the aspect ratio of the rectangle is set to:
\begin{equation}
    \frac{\Delta\theta}{\Delta\phi} = \sin\theta_i,
\end{equation}
which yields:
\begin{align}
    \Delta\theta &= \sqrt{2\pi^2 \frac{N}{HW} p(\omegai)}\cdot \sin\theta_i ,\\
    \Delta\phi &= \sqrt{2\pi^2 \frac{N}{HW} p(\omegai)}.
\end{align}


\section{BSDF Neural Network Parameterization} \label{appendix:parameterization}

Once we have sampled the incoming light directions $\omegai$ and their respective values $L(\point, \omegai)$, we transform them into the local shading frame to calculate the value of the neural shading network $h$. We parameterize the neural network with 2 hidden layers of width 64 as $h(\point, \omegao, \omegai, \uvn)$, where $\point$ is the position, $\omegao, \omegai$ are the outgoing and incoming light directions, respectively, and $\uvn$ is the normal. However, rather than feeding $\omegao$ and $\omegai$ to the network directly, we follow the schema laid out by Rusinkiewicz~\cite{rusinkiewicz1998new} and parameterize the input using the halfway vector $\boldsymbol{\hat{h}}$ and difference vector $\boldsymbol{\hat{d}}$ within the local shading frame $F(\uvn)$, which takes the world space to a frame of reference in which the normal vector points upwards:
\begin{align}
    T&={[0, 0, 1]}^\top \times \uvn \\
    F(\uvn) &= \begin{bmatrix}
    T, & \uvn\times T, & \uvn
    \end{bmatrix}^\top \\
    \boldsymbol{\hat{h}} &= F(\uvn) \frac{\omegai+\omegao}{\|\omegai+\omegao\|_2} \\
    \boldsymbol{\hat{d}} &= F(\boldsymbol{\hat{h}}) \omegai
\end{align}
where $\times$ is the cross product. Finally, we encode these two directions using spherical harmonics up to degree 4 (as done in Ref-NeRF~\cite{verbin2021ref} for encoding view directions), concatenate the feature vector $\feat$ from the field at point $\point$, and pass this as input to the network $h$.

\section{Optimization and Architecture}\label{appendix:optimization}

To calculate the normal vectors of the density field, we apply a finite difference kernel, convolved with a $3\times 3$ Gaussian smoothing kernel with $\sigma=1$, then linearly interpolate between samples to get the resulting gradient in the 3D volume.
We supervise our method using photometric loss, along with the orientation loss of Equation~\ref{eqn:normal_penalty}. Like TensoRF, we use a learning rate of $0.02$ for the rank $1$ and $2$ tensor components, and a learning rate of $10^{-3}$ for everything else. We use Adam~\cite{kingma2014adam} with $\beta_1=0.9,\beta_2=0.99, \varepsilon=10^{-15}$. 
Similar to Ref-NeRF~\cite{verbin2021ref}, we use log-linear learning rate decay with a total decay of $d_w = 10^{-3}$ and a warmup of $N_w = 100$ steps and a decay multiplier of $m_w = 0.1$ over $N_T=3\cdot 10^4$ total iterations. This gives us the following formula for the learning rate multiplier for some iteration $i$:
\begin{equation}
    \left[m_w + (1-m_w) \sin\frac\pi2 \text{clip}\left(\frac{i}{N_w}, 0, 1\right)\right] e^{i/N_T}\log(d_w)
\end{equation}

We initialize the environment map to a constant value of $0.5$. 
Finally, we upsample the resolution of TensoRF from $32^3$ up to $300^3$ cube-root-linearly at steps $500, 1000, 2000, 3000, 4000, 5500, 7000$, and don't shrink the volume to fit the model.

To further reduce the variance of the estimated value of the rendering equation (see Equation~\ref{eqn:montecarlo}), we use quasi-random sampling sequences. Specifically, we use a Sobol sequence~\cite{sable1967} with Owens scrambling~\cite{owen1995randomly}, which gives the procedural sequence necessary for assigning an arbitrary number of secondary ray samples to each primary ray sample. We then apply Cranley-Patterson rotation~\cite{cranley1976randomization} to avoid needing to redraw samples.

\section{Additional Results}
\label{appendix:results}
Tables 2-5 contain full per-scene metrics for our method as well as ablations and baselines. Visual comparisons are also provided in Figures 7-17.

\begin{table*}[]
\resizebox{\linewidth}{!}{
\begin{tabular}{l|rrrrrr|rrrrrrrr}
\rowcolor[HTML]{FFFFFF} 
PSNR $\uparrow$                             & \multicolumn{1}{l}{\cellcolor[HTML]{FFFFFF}teapot} & \multicolumn{1}{l}{\cellcolor[HTML]{FFFFFF}toaster} & \multicolumn{1}{l}{\cellcolor[HTML]{FFFFFF}car} & \multicolumn{1}{l}{\cellcolor[HTML]{FFFFFF}ball} & \multicolumn{1}{l}{\cellcolor[HTML]{FFFFFF}coffee} & \multicolumn{1}{l|}{\cellcolor[HTML]{FFFFFF}helmet} & \multicolumn{1}{l}{\cellcolor[HTML]{FFFFFF}chair} & \multicolumn{1}{l}{\cellcolor[HTML]{FFFFFF}lego} & \multicolumn{1}{l}{\cellcolor[HTML]{FFFFFF}materials} & \multicolumn{1}{l}{\cellcolor[HTML]{FFFFFF}mic} & \multicolumn{1}{l}{\cellcolor[HTML]{FFFFFF}hotdog} & \multicolumn{1}{l}{\cellcolor[HTML]{FFFFFF}ficus} & \multicolumn{1}{l}{\cellcolor[HTML]{FFFFFF}drums} & \multicolumn{1}{l}{\cellcolor[HTML]{FFFFFF}ship} \\ \hline
\rowcolor[HTML]{FFFFFF} 
PhySG$^1$                                   & 35.83                                              & 18.59                                               & 24.40                                           & 27.24                                            & 23.71                                              & 27.51                                               & 21.87                                             & 17.10                                            & 18.02                                                 & 19.16                                           & 24.49                                              & 15.25                                             & 14.35                                             & 18.06                                            \\
\rowcolor[HTML]{FFFFFF} 
NVDiffRec$^1$                               & 40.13                                              & 24.10                                               & 27.13                                           & 30.77                                            & 30.58                                              & 26.66                                               & 32.03                                             & 29.07                                            & 25.03                                                 & 30.72                                           & 33.05                                              & \cellcolor[HTML]{FFD9B3}31.18                     & 24.53                                             & 24.68                                            \\
\rowcolor[HTML]{FFFFFF} 
NVDiffRecMC$^1$                             & 37.91                                              & 21.93                                               & 25.84                                           & 28.89                                            & 29.06                                              & 25.57                                               & 28.13                                             & 26.46                                            & 25.64                                                 & 29.03                                           & 30.56                                              & 25.32                                             & 22.78                                             & 18.59                                            \\
\rowcolor[HTML]{FFB3B3} 
\cellcolor[HTML]{FFFFFF}Ref-NeRF$^2$        & 47.90                                              & \cellcolor[HTML]{FFFFFF}25.70                       & 30.82                                           & 47.46                                            & 34.21                                              & \cellcolor[HTML]{FFFFB4}29.68                       & 35.83                                             & 36.25                                            & 35.41                                                 & 36.76                                           & 37.72                                              & 33.91                                             & 25.79                                             & 30.28                                            \\
\rowcolor[HTML]{FFFFFF} 
Ours, no integral image                     & 42.61                                              & 18.36                                               & 25.32                                           & 21.70                                            & 31.15                                              & 24.82                                               & 30.35                                             & 30.16                                            & 25.62                                                 & 30.03                                           & 33.34                                              & 28.44                                             & 24.04                                             & 25.78                                            \\
\rowcolor[HTML]{FFFFFF} 
Ours, analytical derivative                 & 43.57                                              & 21.57                                               & 27.72                                           & 22.75                                            & 31.08                                              & 28.61                                               & 30.49                                             & 30.23                                            & 28.70                                                 & 31.19                                           & 33.55                                              & 27.83                                             & 24.15                                             & 25.40                                            \\
\rowcolor[HTML]{FFFFB4} 
\cellcolor[HTML]{FFFFFF}Ours, single bounce & 45.23                                              & \cellcolor[HTML]{FFD9B3}26.91                       & 30.13                                           & 38.38                                            & 31.39                                              & \cellcolor[HTML]{FFD9B3}34.32                       & \cellcolor[HTML]{FFD9B3}32.57                     & 32.83                                            & 30.92                                                 & \cellcolor[HTML]{FFD9B3}32.49                   & 35.07                                              & 29.24                                             & \cellcolor[HTML]{FFD9B3}24.99                     & 27.32                                            \\
\rowcolor[HTML]{FFFFFF} 
Ours, no neural                             & 45.21                                              & \cellcolor[HTML]{FFFFB4}25.73                       & 29.03                                           & 37.41                                            & 30.99                                              & 29.63                                               & 30.62                                             & 31.00                                            & 29.37                                                 & 31.29                                           & 33.88                                              & 28.10                                             & 24.52                                             & 26.44                                            \\
\rowcolor[HTML]{FFD9B3} 
\cellcolor[HTML]{FFFFFF}Ours                & 45.29                                              & \cellcolor[HTML]{FFB3B3}27.52                       & 30.28                                           & 38.41                                            & 31.47                                              & \cellcolor[HTML]{FFB3B3}34.38                       & \cellcolor[HTML]{FFFFB4}32.27                     & 32.98                                            & 31.19                                                 & \cellcolor[HTML]{FFFFB4}32.41                   & 35.23                                              & \cellcolor[HTML]{FFFFFF}29.24                     & \cellcolor[HTML]{FFFFB4}24.96                     & 27.37                                           
\end{tabular}
}
\small{$^1$ requires object masks during training. ~~ $^2$ view synthesis method, not inverse rendering. ~~ Red is best, followed by orange, then yellow. }
\caption{\textbf{PSNR Results on the \emph{Shiny Blender} dataset from Ref-NeRF \cite{verbin2021ref} and \emph{Blender} dataset from NeRF \cite{mildenhall2021nerf}.}}
\label{tab:psnrs}
\end{table*}

\begin{table*}[]
\resizebox{\linewidth}{!}{
\begin{tabular}{l|rrrrrr|rrrrrrrr}
\rowcolor[HTML]{FFFFFF} 
SSIM $\uparrow$                             & \multicolumn{1}{l}{\cellcolor[HTML]{FFFFFF}teapot} & \multicolumn{1}{l}{\cellcolor[HTML]{FFFFFF}toaster} & \multicolumn{1}{l}{\cellcolor[HTML]{FFFFFF}car} & \multicolumn{1}{l}{\cellcolor[HTML]{FFFFFF}ball} & \multicolumn{1}{l}{\cellcolor[HTML]{FFFFFF}coffee} & \multicolumn{1}{l|}{\cellcolor[HTML]{FFFFFF}helmet} & \multicolumn{1}{l}{\cellcolor[HTML]{FFFFFF}chair} & \multicolumn{1}{l}{\cellcolor[HTML]{FFFFFF}lego} & \multicolumn{1}{l}{\cellcolor[HTML]{FFFFFF}materials} & \multicolumn{1}{l}{\cellcolor[HTML]{FFFFFF}mic} & \multicolumn{1}{l}{\cellcolor[HTML]{FFFFFF}hotdog} & \multicolumn{1}{l}{\cellcolor[HTML]{FFFFFF}ficus} & \multicolumn{1}{l}{\cellcolor[HTML]{FFFFFF}drums} & \multicolumn{1}{l}{\cellcolor[HTML]{FFFFFF}ship} \\ \hline
\rowcolor[HTML]{FFFFFF} 
PhySG$^1$                                   & .990                                               & .805                                                & .910                                            & .947                                             & .922                                               & .953                                                & .890                                              & .812                                             & .837                                                  & .904                                            & .894                                               & .861                                              & .823                                              & .756                                             \\
\rowcolor[HTML]{FFFFFF} 
NVDiffRec$^1$                               & .993                                               & .898                                                & .938                                            & .949                                             & .959                                               & .931                                                & \cellcolor[HTML]{FFD9B3}.969                      & .949                                             & .923                                                  & \cellcolor[HTML]{FFFFB4}.977                    & \cellcolor[HTML]{FFD9B3}.973                       & \cellcolor[HTML]{FFD9B3}.970                      & .916                                              & \cellcolor[HTML]{FFFFB4}.833                     \\
\rowcolor[HTML]{FFFFFF} 
NVDiffRecMC$^1$                             & .990                                               & .842                                                & .913                                            & .849                                             & .942                                               & .877                                                & .932                                              & .909                                             & .911                                                  & .961                                            & .945                                               & .937                                              & .906                                              & .732                                             \\
\rowcolor[HTML]{FFB3B3} 
\cellcolor[HTML]{FFFFFF}Ref-NeRF$^2$        & .998                                               & .922                                                & .955                                            & .995                                             & .974                                               & \cellcolor[HTML]{FFFFB4}.958                        & .984                                              & .981                                             & .983                                                  & .992                                            & .984                                               & .983                                              & .937                                              & .880                                             \\
\rowcolor[HTML]{FFFFFF} 
Ours, no integral image                     & .994                                               & .734                                                & .895                                            & .753                                             & .959                                               & .880                                                & .946                                              & .946                                             & .896                                                  & .962                                            & .954                                               & .953                                              & .905                                              & .794                                             \\
\rowcolor[HTML]{FFFFFF} 
Ours, analytical derivative                 & \cellcolor[HTML]{FFFFB4}.995                       & .798                                                & .925                                            & .790                                             & .959                                               & .930                                                & .948                                              & .943                                             & .936                                                  & .972                                            & .958                                               & .950                                              & .910                                              & .787                                             \\
\cellcolor[HTML]{FFFFFF}Ours, single bounce & \cellcolor[HTML]{FFD9B3}.996                       & \cellcolor[HTML]{FFFFB4}.909                        & \cellcolor[HTML]{FFD9B3}.951                    & \cellcolor[HTML]{FFD9B3}.983                     & \cellcolor[HTML]{FFD9B3}.962                       & \cellcolor[HTML]{FFB3B3}.971                        & \cellcolor[HTML]{FFFFB4}.964                      & \cellcolor[HTML]{FFD9B3}.966                     & \cellcolor[HTML]{FFFFB4}.957                          & \cellcolor[HTML]{FFD9B3}.978                    & \cellcolor[HTML]{FFFFB4}.969                       & \cellcolor[HTML]{FFFFB4}.959                      & \cellcolor[HTML]{FFD9B3}.922                      & \cellcolor[HTML]{FFD9B3}.835                     \\
\rowcolor[HTML]{FFFFFF} 
Ours, no neural                             & \cellcolor[HTML]{FFD9B3}.996                       & .903                                                & \cellcolor[HTML]{FFFFB4}.945                                            & \cellcolor[HTML]{FFFFB4}.980                                             & .959                                               & .947                                                & .949                                              & .952                                             & .945                                                  & .972                                            & .960                                               & .954                                              & .916                                              & .816                                             \\
\cellcolor[HTML]{FFFFFF}Ours                & \cellcolor[HTML]{FFD9B3}.996                       & \cellcolor[HTML]{FFD9B3}.917                        & \cellcolor[HTML]{FFD9B3}.951                    & \cellcolor[HTML]{FFD9B3}.983                     & \cellcolor[HTML]{FFFFB4}.960                       & \cellcolor[HTML]{FFD9B3}.969                        & \cellcolor[HTML]{FFFFFF}.956                      & \cellcolor[HTML]{FFFFB4}.963                     & \cellcolor[HTML]{FFD9B3}.959                          & \cellcolor[HTML]{FFFFFF}.977                    & \cellcolor[HTML]{FFFFFF}.964                       & \cellcolor[HTML]{FFFFFF}.952                      & \cellcolor[HTML]{FFFFB4}.917                      & \cellcolor[HTML]{FFFFFF}.828                    
\end{tabular}
}
\small{$^1$ requires object masks during training. ~~ $^2$ view synthesis method, not inverse rendering.~~ Red is best, followed by orange, then yellow. }
\caption{\textbf{SSIM Results on the \emph{Shiny Blender} dataset from Ref-NeRF \cite{verbin2021ref} and \emph{Blender} dataset from NeRF \cite{mildenhall2021nerf}.}}
\label{tab:ssims}
\end{table*}

\begin{table*}[]
\resizebox{\linewidth}{!}{
\begin{tabular}{l|rrrrrr|rrrrrrrr}
LPIPS $\downarrow$                          & \multicolumn{1}{l}{teapot}   & \multicolumn{1}{l}{toaster}  & \multicolumn{1}{l}{car}      & \multicolumn{1}{l}{ball}     & \multicolumn{1}{l}{coffee}   & \multicolumn{1}{l|}{helmet}  & \multicolumn{1}{l}{chair}    & \multicolumn{1}{l}{lego}     & \multicolumn{1}{l}{materials} & \multicolumn{1}{l}{mic}      & \multicolumn{1}{l}{hotdog}   & \multicolumn{1}{l}{ficus}    & \multicolumn{1}{l}{drums}    & \multicolumn{1}{l}{ship}     \\ \hline
\rowcolor[HTML]{FFFFFF} 
PhySG$^1$                                   & .022                         & .194                         & .091                         & .179                         & .150                         & .089                         & .122                         & .208                         & .182                          & .108                         & .163                         & .144                         & .188                         & .343                         \\
\rowcolor[HTML]{FFFFFF} 
NVDiffRec$^1$                               & .022                         & .180                         & .057                         & .194                         & .097                         & .134                         & \cellcolor[HTML]{FFD9B3}.027 & .037                         & .104                          & .033                         & \cellcolor[HTML]{FFFFB4}.038 & \cellcolor[HTML]{FFD9B3}.030 & .070                         & .208                         \\
\rowcolor[HTML]{FFFFFF} 
NVDiffRecMC$^1$                             & .029                         & .243                         & .086                         & .346                         & .131                         & .215                         & .080                         & .075                         & .096                          & .057                         & .089                         & .076                         & .096                         & .319                         \\
\rowcolor[HTML]{FFB3B3} 
\cellcolor[HTML]{FFFFFF}Ref-NeRF$^2$        & .004                         & .095                         & \cellcolor[HTML]{FFFFFF}.041 & \cellcolor[HTML]{FFFFFF}.059 & \cellcolor[HTML]{FFFFFF}.078 & \cellcolor[HTML]{FFFFB4}.075 & .017                         & \cellcolor[HTML]{FFB3B3}.018 & .022                          & .007                         & .022                         & .019                         & .059                         & \cellcolor[HTML]{FFD9B3}.139 \\
\rowcolor[HTML]{FFFFFF} 
Ours, no integral image                     & .013                         & .285                         & .077                         & .399                         & \cellcolor[HTML]{FFD9B3}.065 & .180                         & .055                         & .031                         & .074                          & .042                         & .051                         & .039                         & .077                         & .180                         \\
\rowcolor[HTML]{FFFFFF} 
Ours, analytical derivative                 & .011                         & .235                         & .053                         & .353                         & .071                         & .118                         & .052                         & .031                         & .048                          & .028                         & .047                         & .043                         & .075                         & .190                         \\
\cellcolor[HTML]{FFFFFF}Ours, single bounce & \cellcolor[HTML]{FFD9B3}.008 & \cellcolor[HTML]{FFFFB4}.114 & \cellcolor[HTML]{FFB3B3}.033 & \cellcolor[HTML]{FFD9B3}.047 & \cellcolor[HTML]{FFB3B3}.063 & \cellcolor[HTML]{FFB3B3}.050 & \cellcolor[HTML]{FFFFB4}.032 & \cellcolor[HTML]{FFB3B3}.018 & \cellcolor[HTML]{FFD9B3}.026  & \cellcolor[HTML]{FFD9B3}.020 & \cellcolor[HTML]{FFD9B3}.034 & \cellcolor[HTML]{FFFFB4}.033 & \cellcolor[HTML]{FFD9B3}.065 & \cellcolor[HTML]{FFB3B3}.135 \\
\rowcolor[HTML]{FFFFFF} 
Ours, no neural                             & \cellcolor[HTML]{FFD9B3}.008 & .115                         & \cellcolor[HTML]{FFFFB4}.039 & \cellcolor[HTML]{FFFFB4}.058 & .071                         & .090                         & .053                         & \cellcolor[HTML]{FFD9B3}.026 & \cellcolor[HTML]{FFFFB4}.036  & .027                         & .045                         & .036                         & .070                         & .161                         \\
\cellcolor[HTML]{FFFFFF}Ours                & \cellcolor[HTML]{FFFFB4}.010 & \cellcolor[HTML]{FFD9B3}.104 & \cellcolor[HTML]{FFD9B3}.034 & \cellcolor[HTML]{FFB3B3}.046 & \cellcolor[HTML]{FFFFB4}.069 & \cellcolor[HTML]{FFD9B3}.055 & \cellcolor[HTML]{FFFFFF}.044 & \cellcolor[HTML]{FFFFB4}.024 & \cellcolor[HTML]{FFD9B3}.026  & \cellcolor[HTML]{FFFFB4}.022 & \cellcolor[HTML]{FFFFFF}.046 & \cellcolor[HTML]{FFFFFF}.044 & \cellcolor[HTML]{FFFFB4}.068 & \cellcolor[HTML]{FFFFB4}.149
\end{tabular}
}
\small{$^1$ requires object masks during training. ~~ $^2$ view synthesis method, not inverse rendering. ~~Red is best, followed by orange, then yellow. }
\caption{\textbf{LPIPS Results on the \emph{Shiny Blender} dataset from Ref-NeRF \cite{verbin2021ref} and \emph{Blender} dataset from NeRF \cite{mildenhall2021nerf}.}}
\label{tab:lpips}
\end{table*}

\begin{table*}[]
\resizebox{\linewidth}{!}{
\begin{tabular}{
>{\columncolor[HTML]{FFFFFF}}l |
>{\columncolor[HTML]{FFFFFF}}r 
>{\columncolor[HTML]{FFFFFF}}r 
>{\columncolor[HTML]{FFFFFF}}r 
>{\columncolor[HTML]{FFFFFF}}r 
>{\columncolor[HTML]{FFFFFF}}r 
>{\columncolor[HTML]{FFFFFF}}r |
>{\columncolor[HTML]{FFFFFF}}r 
>{\columncolor[HTML]{FFFFFF}}r 
>{\columncolor[HTML]{FFFFFF}}r 
>{\columncolor[HTML]{FFFFFF}}r 
>{\columncolor[HTML]{FFFFFF}}r 
>{\columncolor[HTML]{FFFFFF}}r 
>{\columncolor[HTML]{FFFFFF}}r 
>{\columncolor[HTML]{FFFFFF}}r }
$\text{MAE}^\circ \downarrow$ & \multicolumn{1}{l}{\cellcolor[HTML]{FFFFFF}teapot} & \multicolumn{1}{l}{\cellcolor[HTML]{FFFFFF}toaster} & \multicolumn{1}{l}{\cellcolor[HTML]{FFFFFF}car} & \multicolumn{1}{l}{\cellcolor[HTML]{FFFFFF}ball} & \multicolumn{1}{l}{\cellcolor[HTML]{FFFFFF}coffee} & \multicolumn{1}{l|}{\cellcolor[HTML]{FFFFFF}helmet} & \multicolumn{1}{l}{\cellcolor[HTML]{FFFFFF}chair} & \multicolumn{1}{l}{\cellcolor[HTML]{FFFFFF}lego} & \multicolumn{1}{l}{\cellcolor[HTML]{FFFFFF}materials} & \multicolumn{1}{l}{\cellcolor[HTML]{FFFFFF}mic} & \multicolumn{1}{l}{\cellcolor[HTML]{FFFFFF}hotdog} & \multicolumn{1}{l}{\cellcolor[HTML]{FFFFFF}ficus} & \multicolumn{1}{l}{\cellcolor[HTML]{FFFFFF}drums} & \multicolumn{1}{l}{\cellcolor[HTML]{FFFFFF}ship} \\ \hline
PhySG$^1$                     & 6.634                                              & \cellcolor[HTML]{FFFFB4}9.749                       & 8.844                                           & \cellcolor[HTML]{FFB3B3}0.700                    & 22.514                                             & \cellcolor[HTML]{FFB3B3}2.324                       & 18.569                                            & 40.244                                           & 18.986                                                & 26.053                                          & 28.572                                             & 35.974                                            & \cellcolor[HTML]{FFD9B3}21.696                    & 43.265                                           \\
NVDiffRec$^1$                 & \cellcolor[HTML]{FFB3B3}3.874                      & 14.336                                              & 15.286                                          & 5.584                                            & \cellcolor[HTML]{FFB3B3}11.132                     & 20.513                                              & 25.023                                            & 42.978                                           & 26.969                                                & 26.571                                          & 29.115                                             & 38.647                                            & 26.512                                            & 39.262                                           \\
NVDiffRecMC$^1$               & 5.928                                              & 11.905                                              & \cellcolor[HTML]{FFFFB4}8.357                   & 1.313                                            & 18.385                                             & 8.131                                               & 23.469                                            & 42.706                                           & 9.132                                                 & 26.184                                          & 26.470                                             & \cellcolor[HTML]{FFB3B3}34.324                    & 25.219                                            & 41.952                                           \\
Ref-NeRF$^2$                  & 9.234                                              & 42.870                                              & 14.927                                          & 1.548                                            & \cellcolor[HTML]{FFD9B3}12.240                     & 29.484                                              & 19.852                                            & \cellcolor[HTML]{FFB3B3}24.469                   & 9.531                                                 & 24.938                                          & \cellcolor[HTML]{FFFFB4}13.211                     & 41.052                                            & 27.853                                            & 31.707                                           \\
Ours, no integral image       & 10.078                                             & 39.779                                              & 28.744                                          & 45.998                                           & 14.776                                             & 28.550                                              & 20.594                                            & 26.712                                           & 27.462                                                & 29.956                                          & 15.188                                             & 36.543                                            & 32.118                                            & 37.464                                           \\
Ours, analytical derivative   & 6.400                                              & 21.403                                              & 10.685                                          & 21.145                                           & 15.425                                             & 8.800                                               & 17.801                                            & 26.852                                           & \cellcolor[HTML]{FFFFB4}8.960                         & \cellcolor[HTML]{FFB3B3}19.426                  & 14.138                                             & \cellcolor[HTML]{FFFFB4}35.505                    & 27.333                                            & 36.423                                           \\
Ours, single bounce           & 6.343                                              & \cellcolor[HTML]{FFD9B3}7.133                       & \cellcolor[HTML]{FFD9B3}7.746                   & \cellcolor[HTML]{FFFFB4}0.722                    & \cellcolor[HTML]{FFFFB4}12.950                     & \cellcolor[HTML]{FFFFB4}2.401                       & \cellcolor[HTML]{FFB3B3}14.285                    & \cellcolor[HTML]{FFFFB4}26.082                   & \cellcolor[HTML]{FFD9B3}8.315                         & \cellcolor[HTML]{FFD9B3}20.004                  & \cellcolor[HTML]{FFD9B3}10.263                     & 37.498                                            & \cellcolor[HTML]{FFFFB4}22.358                    & \cellcolor[HTML]{FFB3B3}29.771                   \\
Ours, no neural               & \cellcolor[HTML]{FFD9B3}4.508                      & 10.288                                              & 8.388                                           & \cellcolor[HTML]{FFD9B3}0.703                    & 14.745                                             & 5.320                                               & \cellcolor[HTML]{FFFFB4}17.503                    & 28.290                                           & 9.549                                                 & 20.181                                          & 13.356                                             & \cellcolor[HTML]{FFD9B3}35.298                    & 24.651                                            & \cellcolor[HTML]{FFFFB4}30.326                   \\
Ours                          & \cellcolor[HTML]{FFFFB4}5.672                      & \cellcolor[HTML]{FFB3B3}6.660                       & \cellcolor[HTML]{FFB3B3}7.742                   & 0.723                                            & 13.173                                             & \cellcolor[HTML]{FFD9B3}2.395                       & \cellcolor[HTML]{FFD9B3}14.330                    & \cellcolor[HTML]{FFD9B3}25.918                   & \cellcolor[HTML]{FFB3B3}8.101                         & \cellcolor[HTML]{FFFFB4}20.144                  & \cellcolor[HTML]{FFB3B3}10.043                     & 37.405                                            & \cellcolor[HTML]{FFB3B3}21.524                    & \cellcolor[HTML]{FFD9B3}30.152                  
\end{tabular}
}
\small{$^1$ requires object masks during training. ~~ $^2$ view synthesis method, not inverse rendering. ~~Red is best, followed by orange, then yellow. }
\caption{\textbf{MAE Results on the \emph{Shiny Blender} dataset from Ref-NeRF \cite{verbin2021ref} and \emph{Blender} dataset from NeRF \cite{mildenhall2021nerf}.}}
\label{tab:maes}
\end{table*}

\plotscene{ball} 



\plotscene{coffee}

\plotscenetop{teapot} 

\plotscene{toaster} 

\plotsceneshort{materials} 

\plotscene{drums} 

\plotscene{ficus} 

\plotscenebottom{hotdog} 

\plotscene{mic} 

\plotship{ship}

\plotscene{lego} 

\end{document}